\documentclass{article}

\usepackage{preprint}

\title{BoTier: Multi-Objective Bayesian Optimization \\ with Tiered Composite Objectives}

\begin{document}

%% Math Utils

\newcommand{\EX}[2]{\mathbb{E}\left[#1 \mid #2 \right]}
\newcommand{\argmax}[2]{\underset{#1}{\text{argmax}} \ #2}

% Misc Utils

\newcommand{\etal}{\textit{et al.}}

\maketitle

\begin{authors}
Mohammad Haddadnia,$^{a, b}$ Leonie Grashoff,$^c$ and Felix Strieth-Kalthoff$^{~c, d,\ast}$
\end{authors}

\begin{affiliations}
\item[$^a$]{Harvard University, Department of Biological Chemistry \& Molecular Pharmacology, Boston (MA), United States.}
\item[$^b$]{Harvard University, Dana-Farber Cancer Institute, Department of Cancer Biology, Boston (MA), United States.}
\item[$^c$]{University of Wuppertal, School of Mathematics and Natural Sciences, Wuppertal, Germany.}
\item[$^d$]{University of Wuppertal, Interdisciplinary Center for Machine Learning and Data Analytics, Wuppertal, Germany.}
\item[$^{\ast}$]{E-Mail: strieth-kalthoff@uni-wuppertal.de}
\end{affiliations}

\begin{abstract}
Scientific optimization problems are usually concerned with balancing multiple competing objectives, which come as preferences over both the outcomes of an experiment (e.g. maximize the reaction yield) and the corresponding input parameters (e.g. minimize the use of an expensive reagent). 
Typically, practical and economic considerations define a hierarchy over these objectives, which must be reflected in algorithms for sample-efficient experiment planning. 
Herein, we introduce \textit{BoTier}, a composite objective that can flexibly represent a hierarchy of preferences over both experiment outcomes and input parameters. 
We provide systematic benchmarks on synthetic and real-life surfaces, demonstrating the robust applicability of \textit{BoTier} across a number of use cases. 
Importantly, \textit{BoTier} is implemented in an auto-differentiable fashion, enabling seamless integration with the \textit{BoTorch} library, thereby facilitating adoption by the scientific community. 
\end{abstract}

\section{Introduction}

Multi-objective optimization (MOO) – the task of finding a global optimum that simultaneously satisifies a set of optimization criteria – is a common problem in many fields of science and engineering.\cite{Fromer2023, Vel2024, Shields2021, Angello2022, Shi2023} 
As an example, a new drug needs to simultaneously optimize target activity, side effects, bioavailability and metabolic profile; similarly, a new material must meet several demands relating to properties, stability or synthesizability. 
Usually, such objectives are conflicting, so the \textit{optimal} solution represents a trade-off between them. 
In many scenarios, finding these optimal solutions is cumbersome, especially in a setting in which experimental evaluations are expensive – creating a need for efficient experiment planning algorithms. Over the past decade, Bayesian Optimization (BO) has become the \textit{de-facto} choice for sample-efficient iterative optimization of black-box functions,\cite{Snoek2012, Frazier2018, Garnett2023} and has found particular popularity in the context of autonomous experimentation with self-driving laboratories (SDLs).\citep{Tom2024, StriethKalthoff2024}

\begin{figure}[h]    
 \centering
 \includegraphics[width=9cm]{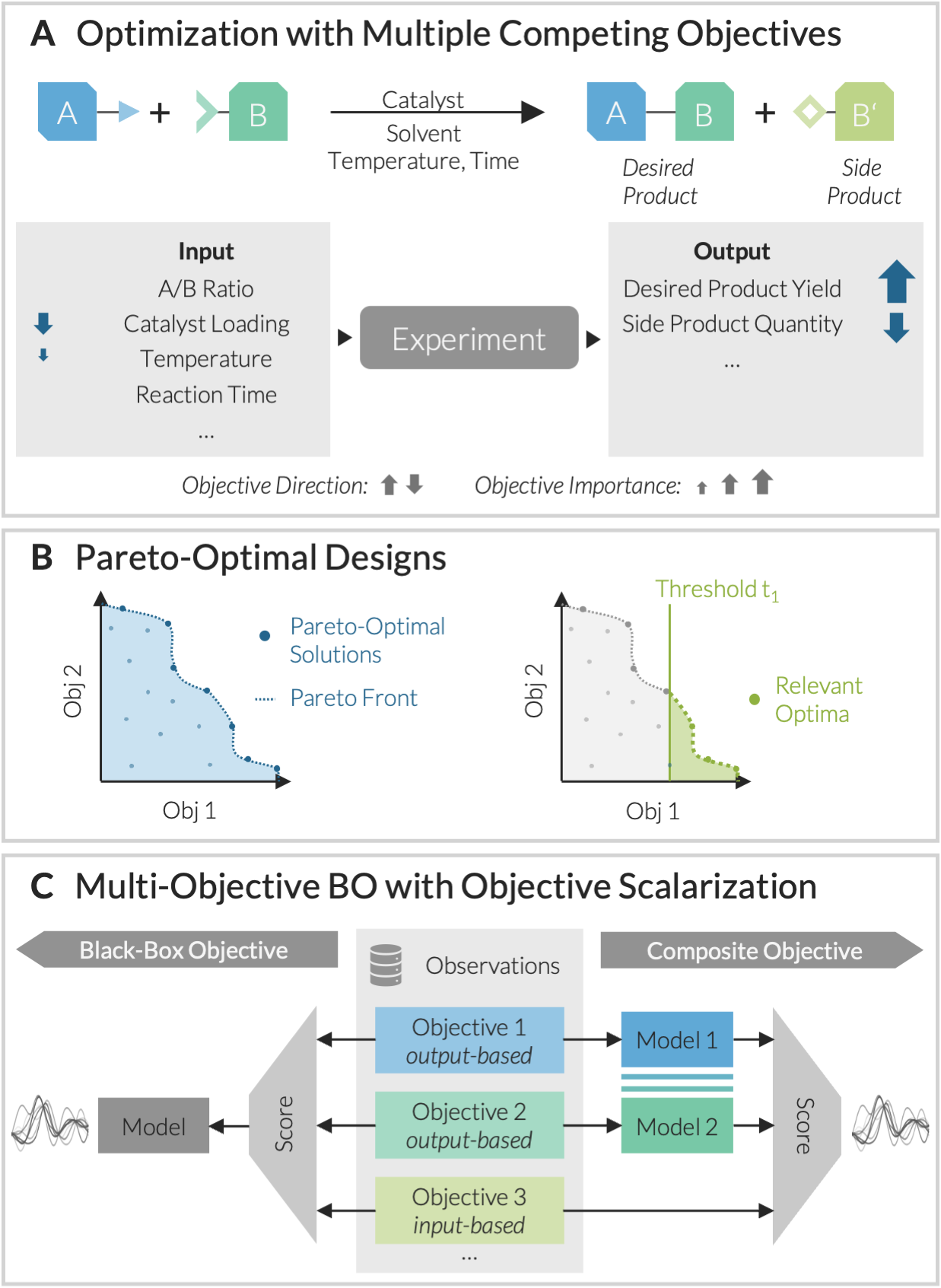}
 \caption{MOO with preferences over experiment inputs and outputs. A) Example from chemical reaction optimization. B) Pareto front for two competing objectives. C) Workflow of BO with multi-objective scalarization in a \textit{black-box}  (left) and \textit{composite} manner (right, this work).}
 \label{fig:intro}
\end{figure}

In scientific optimization problems, the primary objective(s) are generally derived from the outcome of an experiment. 
In reaction optimization (Fig. \ref{fig:intro}a), for example, this could be the yield of the desired product, or the quantity of an undesired side product. 
At the same time, secondary optimization objectives can include preferences over input parameters, such as minimizing the loading of an expensive catalyst, or minimizing the reaction temperature to lower energy consumption.\cite{Clayton2019, Torres2022, Taylor2023}
It is worth noting that such considerations imply that certain objectives are prioritized over others, establishing a known hierarchy.\cite{Waltz1967, Stadler1988, Rentmeesters1996}

In MOO, a solution in which further improving one objective is detrimental for at least one other objective, is called a \textit{Pareto optimum},\cite{Deb2011} and the set of all Pareto optima is referred to as the \textit{Pareto front} (Fig. \ref{fig:intro}b).
In an ideal scenario, knowing the entire Pareto front would enable optimal \textit{post-hoc} decisions on the desired optimum, accounting for all inter-objective trade-offs.
Accordingly, the past decades have seen significant advances in hypervolume-based acquisition functions to map the Pareto front.\cite{Emmerich2006, Daulton2020}
At the same time, Pareto-oriented optimization may spend significant experimental resources on mapping regions of the Pareto front that are not of interest to the researcher (Fig. \ref{fig:intro}b). Therefore, when objective importances are known, scalarizing multiple objectives into a single score can help guide the optimization to desired regions of the Pareto front.\cite{Chugh2020, Fromer2023, Taylor2023}

In practice, such scalar scores are often used in a "black-box" manner (Fig. \ref{fig:intro}c (left)), where each observation's multiple objective values are first concatenated into a single score, and standard single-objective BO is then employed to optimize this score over the search space.\cite{Chugh2020, Fromer2023}
While straightforward, this approach has two main drawbacks: 
(a) if input-based objectives are included, their known dependence on input parameters must be "re-learned" by the surrogate model, likely reducing optimization efficiency; 
(b) the scalar score itself is artificial and may not carry physical meaning, which can hinder the design of effective priors.\citep{Klarner2024, Kristiadi2024}
To address these issues, Frazier and co-workers introduced the concept of \textit{composite objectives},\citep{Astudillo2019} which apply a scalar score only \textit{after} building surrogate models. Notably, calculating such composite objectives requires operating on (multiple) model posterior distributions, complicating practical implementation. 

Combining the principle of hierarchical MOO with the idea of composite objectives, we herein introduce \textit{BoTier} as a flexible framework for MOO with tiered preferences over both experiment inputs and outputs. The main contributions of this paper include 1) the formulation of an improved, fully auto-differentiable hierarchical scalarization composite objective; 2) its open-source implementation as an extension of the \textit{BoTorch} library; and 3) systematic benchmarks on analytical surfaces and real-world chemistry examples, showcasing how \textit{BoTier} can efficiently navigate MOO problems in the context of scientific optimization. 
\section{Formulation of \textit{BoTier}}

\subsection{Scalarization Functions}

Given a set of $N$ objectives $\{\psi_i \}_{i=1}^N$, a scalarization function combines these objective values into a single score that reflects user preferences. 
In the context of hierarchical optimization, this was pioneered by Aspuru-Guzik and co-workers, who introduced \textit{Chimera} $\chi$ as an additive scalarization function.\cite{Haese2021} 
Here, it is shown for the case where each objective is to be maximized (see eq. \ref{eq:chimera}).
Each objective has a user-defined ``satisfaction threshold'' $t_i$. 
The component-wise formulation of $\chi$ then ensures that the contribution of objective $\psi_i$ to the score is only considered if all superordinate objectives $\{\psi_j \}_{j<i}$ meet their respective thresholds. 
Additionally, the contribution of each objective is shifted by the highest observed values of all superordinate objectives to ensure continuity of $\chi$. 
If all objectives meet their thresholds, the primary objective is used for optimization (third line of eq. \ref{eq:chimera}). 

\begin{equation}
  \begin{split}
    \chi = \ &\psi_1 \cdot H(t_1 - \psi_1) \\
    &+ \sum_{i=2}^N \left( \Big( \psi_i + \sum_{j=1}^{i-1} \max_{x' \in \mathcal{X}_j} \psi_j(x') \Big) \cdot H(t_i - \psi_i) \cdot \prod_{j=1}^{i-1} H(\psi_j - t_j) \right) \\ 
    &+ \left( \psi_1 + \sum_{j=1}^{N} \max_{x' \in \mathcal{X}_j} \psi_j(x') \right) \cdot \prod_{j=1}^{N} H(\psi_j - t_j)
  \end{split}
\label{eq:chimera}
\end{equation}

Here, $H(x)$ is the Heaviside step function. 

Although \textit{Chimera} is widely used for MOO, its current formulation is limited to "black-box objective" scenarios, where $\chi$ is computed for all $K$ observations, $\{(\psi_1(x_k), \psi_2(x_k), ..., \psi_N(x_k))\}_{k=1}^K$. Since the value of $\chi$ for a single observation depends on all other observations considered at the same time, batch-wise evaluation of $\chi$ is not possible in its current form. Moreover, its implementation is not auto-differentiable, limiting its usefulness as a composite objective for BO. 

\subsection{Formulation of \textit{BoTier}}

We address these limitations by proposing an alternative hierarchical scalarization score designed as a composite objective. Without loss of generality, we assume a maximization problem for each objective, and define 

\begin{equation}
    \Xi = \sum_{i=1}^N \left( \min(\psi_i, t_i) \cdot \prod_{j=1}^{i-1} H(\psi_j - t_j) \right)
\label{eq:our_scalarization}
\end{equation}

As in \textit{Chimera}, the product $\prod_{j=1}^{i-1} H(\psi_j - t_j)$ ensures that an objective $\psi_i$ only contributes to $\Xi$ once all superordinate objectives $\psi_j$ ($j<i$) have met their satisfaction thresholds. If $\psi_i$ is below its threshold $t_i$, it becomes the limiting objective in that region of parameter space, and $\min(\psi_i, t_i)$ returns $\psi_i$. Otherwise, it adds $t_i$ to the score, preserving the continuity of $\Xi$. Moreover, this formulation is consistent with the ranking behavior of \textit{Chimera} (see SI, section 4).

All $\psi_i$ can optionally be normalized to the range $[0, 1]$ based on expert knowledge. This normalization places gradients of $\Xi$ on a consistent scale for all $x \in \mathcal{X}$. 

Our implementation of $\Xi$ uses continuously differentiable approximations for both $\min(x_1, x_2)$ and $H(x)$ (see Supplementary Information (SI), section 1), enabling automatic differentiation in the \textit{PyTorch} framework. This makes gradient-based optimization efficient, and integrates seamlessly with the \textit{BoTorch} ecosystem for BO. As a composite objective, $\Xi$ and can be flexibly evaluated over both experiment inputs and model outputs (i.e. posterior distributions) using Monte-Carlo integration (see SI, section 1.3 for further details).\cite{Balandat2020, Daulton2022} We provide \textit{BoTier} as a lightweight Python library, which can be installed from the Python Package Index (PyPI). 
\section{Experimental Use Cases}

We evaluated the applicability and limitations of \textit{BoTier} through benchmark studies on analytical multi-objective test surfaces and real-life optimization problems from chemistry and materials science. Herein, we set out to investigate the influence of two major algorithm choices on sample efficiency in multi-objective BO: a) the use of \textit{BoTier} compared to other MOO strategies. b) the application of composite objectives relative to "black-box" approaches. 

\begin{figure}[h]    
 \centering
 \includegraphics[width=9cm]{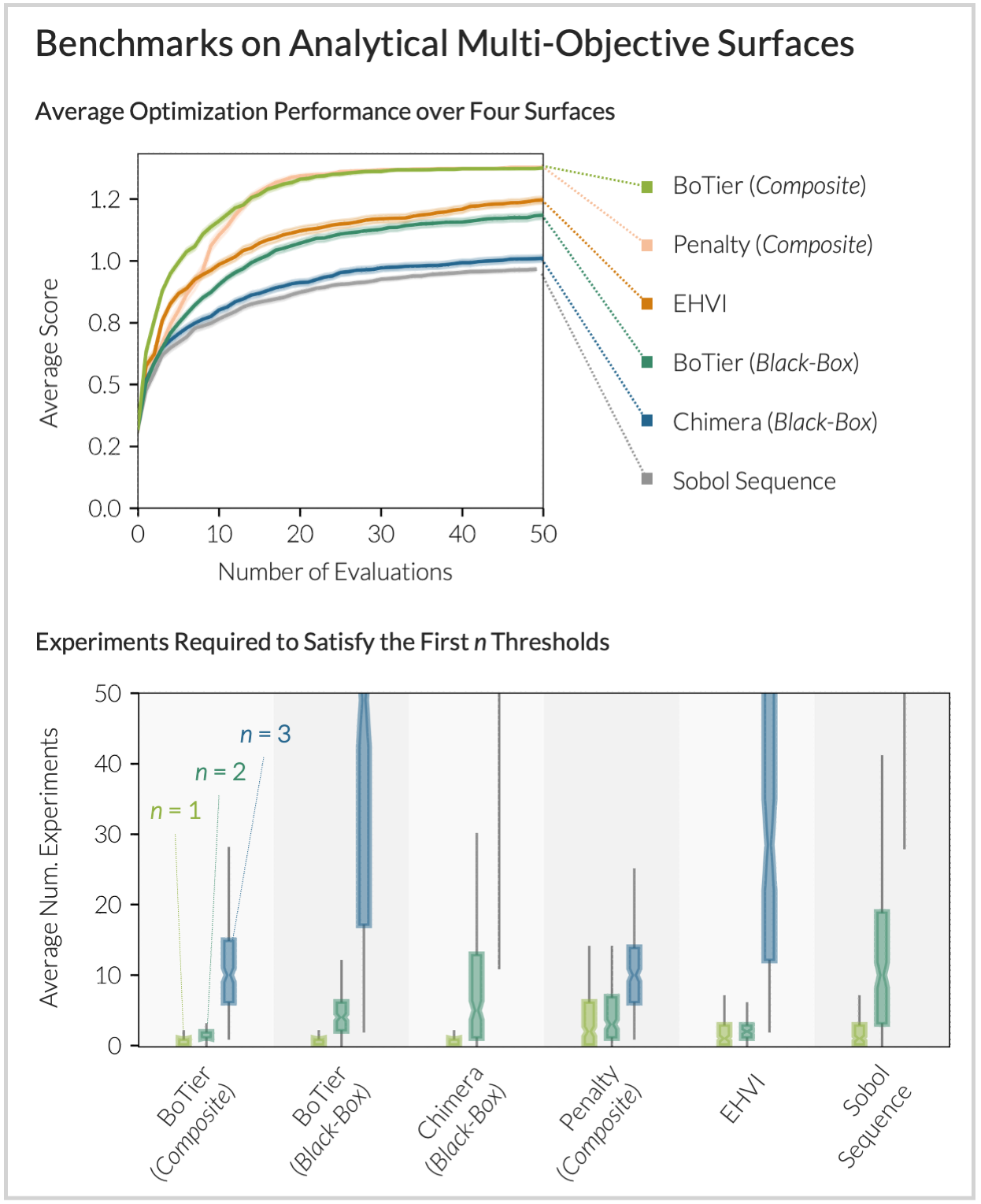}
 \caption{Benchmarks of different MOO strategies on four analytical surfaces, each extended by an input-dependent objective (see SI for further details). Top panel: Best observed value of $\Xi$ as a function of the number of experimental evaluations. All statistics were calculated on 50 independent campaigns on each surface. Intervals are plotted as the standard error. Bottom panel: Number of experiments required to satisfy the first objective ($n=1$, green); the first two objectives ($n=2$, dark green); or all three objectives ($n=3$, blue).}
 \label{fig:analytical_surfaces}
\end{figure}

First, several MOO strategies were evaluated on four analytical surfaces from the \textit{BoTorch} library, augmented by secondary objectives that depend on the function inputs (see SI for further details). 
Across all surfaces, \textit{BoTier}, used as a black-box objective, already converged faster to the optimum value of $\Xi$ than \textit{Chimera} (Fig. \ref{fig:analytical_surfaces}, upper row). Likewise, the number of experiments needed to find conditions that satisfy the first objective, the first two objectives, or all three objectives, was consistently lower (Fig. \ref{fig:analytical_surfaces} bottom row). 
In every case, using \textit{BoTier} as a composite objective further accelerated optimization, which we initially attributed to the the surrogate model not needing to "re-discover" known correlations between inputs and objectives. 
Surprisingly, this finding also held true for those MOO problems in which all objectives solely depended on the experiment outputs (see SI, Section 2.2). 
While a systematic analysis is beyond the scope of this study, these findings suggest that it is simpler to model two distributions independently than a more complex joint distribution (as in the case of black-box objectives). 
Indeed, multi-output GP surrogates, that try to capture cross-correlations between objectives, did not improve optimization performance compared to single-task GP models in most cases (see Figs. S12 and S17). 

Moreover, we evaluated \textit{BoTier} against a threshold-based, but non-hierarchical composite objective: a penalty-based scalarization technique introduced by deMello and co-workers.\citep{Walker2017}. 
The widespread Pareto-oriented Expected Hypervolume Improvement (EHVI) acquisition function was used as a baseline.\cite{Emmerich2011}. 
While optimization behavior varied by problem (see SI, Sections 2.2 and 2.3), several trends emerged: 
Compared to \textit{BoTier}, the penalty-based composite objective often takes more evaluations to satisfy the early objectives in the hierarchy.
The identification of points that satisfy all objective criteria is achieved at a comparable experimental budget, highlighting the general efficiency of composite objectives. 
As expected, EHVI, lacking preferences for any "region" of the Pareto front, required substantially more experiments to identify Pareto-optimal points that satisfy all criteria (see Fig.~\ref{fig:analytical_surfaces} and SI). 
These benchmarks confirm the feasibility of \textit{BoTier}'s formulation and its effectiveness as a composite objective. 

\begin{figure*}[h]    
 \centering
 \includegraphics[width=17cm]{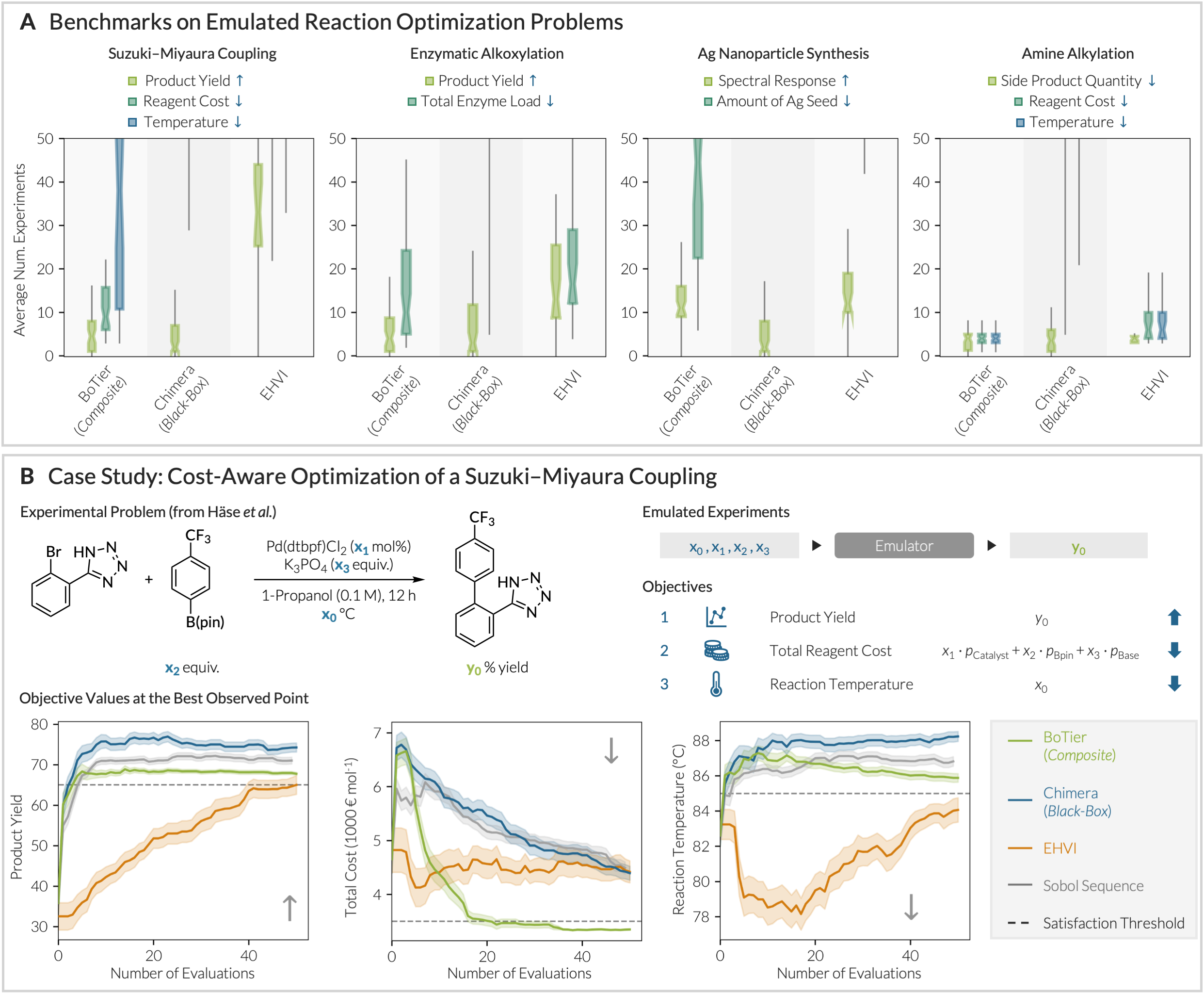}
 \caption{Evaluation of different MOO strategies for chemical reaction optimization. A) Optimization performance on different emulated reaction optimization problems. Number of experiments required to satisfy the first $n$ objectives. B) Case study of a Suzuki-Miyaura coupling. Plots show the trajectories of the respective objective values at the best experimental data point observed so far. See SI for further details.}
 \label{fig:emulated_reactions}
\end{figure*}

Encouraged by these results, we tested \textit{BoTier} on scenarios which are highly relevant to self-driving laboratories. Using emulated experiments, as described by Häse \etal\citep{Haese2018}, MOO algorithms were evaluated for condition optimization in a heterocyclic Suzuki–Miyaura coupling, an enzymatic alkoxylation reaction, a synthesis of silver nanoparticles monitored through spectrophotometry,\citep{mekki-berrada2021nanoparticle} and an amine monoalkylation.\citep{Schweidtmann2018}
Fig. \ref{fig:emulated_reactions}b illustrates the Suzuki–Miyaura coupling, which is optimized over reagent stoichiometry, catalyst loading, base loading and reaction temperature. 
Process chemistry considerations define a three-tier objective hierarchy, consisting of 1) maximized product yield, 2) minimized cost of all reactants and reagents, and 3) minimized reaction temperature.

Using an existing dataset on Suzuki coupling yields, a regression model was trained to predict the yield over the full search space (see Fig. \ref{fig:emulated_reactions}b). 
Benchmarking different MOO algorithms on this problem shows that the \textit{BoTier} composite objective is the only strategy capable of identifying reaction conditions that simultaneously meet all thresholds. 
By contrast, a statistical, model-free baseline (Sobol sampling, see SI for further details) rapidly found high-yielding conditions, but failed to keep cost and temperature low. Similarly, neither black-box \textit{Chimera} nor the evaluation of the full Pareto front (EHVI) identified satisfactory conditions at the budget of 50 optimization experiments.

Similar trends were observed for the other emulated problems (Fig. \ref{fig:emulated_reactions}a).
Notably, we observed cases in which \textit{BoTier} and EHVI satisfy the criteria at similar rates (Fig. \ref{fig:emulated_reactions}a, panels 2 and 4); which occured when the objectives did not strongly compete (see Figs. S21 and S23 for further details). 
Overall, if a hierarchy between objectives exists, \textit{BoTier} proved to be a robust scalarization objective which, in all cases investigated, never performed worse, but often performed notably better than existing MOO methods – particularly when applied as a composite objective. 
\section{Outlook and Perspectives}

We have introduced \textit{BoTier} as a flexible composite objective for hierarchical multi-objective BO. 
Based on our benchmark studies, we formulate the following empirical guidelines for MOO:  
\begin{enumerate}
\item{\textbf{Use hierarchical objectives when a hierarchy exists.}
If the objectives, whether input- and output-dependent, are subject to a well-defined priority structure, \textit{BoTier} offers a robust objective to encode and optimize these preferences. 
Our benchmarks show that, in these cases, hierarchical methods can more rapidly identify desirable optima than approaches that seek to map the entire Pareto front.}
\item{\textbf{Favor composite over black-box objectives whenever possible.}
Across all problems studied, composite objectives consistently outperformed black-box approaches, often yielding substantial speedups. 
In no case did a black-box objective prove superior. 
We foresee that this effect wil be even more pronounced when incorporating priors over physically meaningful quantities.}
\end{enumerate}

To encourage broader adoption, \textit{BoTier} is provided as a lightweight, open-source extension to the \textit{BoTorch} library. 
Looking ahead, we are exploring its applications in self-driving laboratories, where hierarchical optimization can be especially valuable for balancing complex objectives including materials properties, synthetic feasibility, cost, and sustainability. 
We anticipate that \textit{BoTier} will be a valuable addition to the optimization toolbox for autonomous research systems.
\section*{Author Contributions}
Conceptualization: F.S.-K.; 
Data Curation: M.H., L.G., F.S.-K.; 
Formal Analysis: M.H., L.G., F.S.-K.; 
Funding Acquisition: F.S.-K.; 
Investigation: M.H., L.G., F.S.-K.; 
Methodology: M.H., F.S.-K.; 
Project Administration: F.S.-K.; 
Resources: F.S.-K.; 
Software: M.H., F.S.-K.; 
Supervision: F.S.-K.; 
Validation: M. H., L.G., F.S.-K.; 
Visualization: M. H., F.S.-K.; 
Writing – original draft: F.S.-K.; 
Writing – review \& editing: M.H., L.G., F.S.-K.

\section*{Conflicts of interest}
There are no conflicts to declare.

\section*{Data availability}
The software package is available on \textit{Github} under \href{https://github.com/fsk-lab/botier}{https://github.com/fsk-lab/botier}, and can be obtained from the Python Package Index (PyPI) at \href{https://pypi.org/project/botier}{https://pypi.org/project/botier}. 
Scripts for reproducing all results are available on \textit{GitHub}.

\bibliography{literature} 

\providecommand*{\mcitethebibliography}{\thebibliography}
\csname @ifundefined\endcsname{endmcitethebibliography}
{\let\endmcitethebibliography\endthebibliography}{}
\begin{mcitethebibliography}{33}
\providecommand*{\natexlab}[1]{#1}
\providecommand*{\mciteSetBstSublistMode}[1]{}
\providecommand*{\mciteSetBstMaxWidthForm}[2]{}
\providecommand*{\mciteBstWouldAddEndPuncttrue}
  {\def\EndOfBibitem{\unskip.}}
\providecommand*{\mciteBstWouldAddEndPunctfalse}
  {\let\EndOfBibitem\relax}
\providecommand*{\mciteSetBstMidEndSepPunct}[3]{}
\providecommand*{\mciteSetBstSublistLabelBeginEnd}[3]{}
\providecommand*{\EndOfBibitem}{}
\mciteSetBstSublistMode{f}
\mciteSetBstMaxWidthForm{subitem}
{(\emph{\alph{mcitesubitemcount}})}
\mciteSetBstSublistLabelBeginEnd{\mcitemaxwidthsubitemform\space}
{\relax}{\relax}

\bibitem[Fromer and Coley(2023)]{Fromer2023}
J.~C. Fromer and C.~W. Coley, \emph{Patterns}, 2023, \textbf{4}, 100678\relax
\mciteBstWouldAddEndPuncttrue
\mciteSetBstMidEndSepPunct{\mcitedefaultmidpunct}
{\mcitedefaultendpunct}{\mcitedefaultseppunct}\relax
\EndOfBibitem
\bibitem[Vel \emph{et~al.}(2024)Vel, Cortés-Borda, and Felpin]{Vel2024}
A.~S. Vel, D.~Cortés-Borda and F.-X. Felpin, \emph{React. Chem. Eng.}, 2024, \textbf{9}, 2882--2891\relax
\mciteBstWouldAddEndPuncttrue
\mciteSetBstMidEndSepPunct{\mcitedefaultmidpunct}
{\mcitedefaultendpunct}{\mcitedefaultseppunct}\relax
\EndOfBibitem
\bibitem[Shields \emph{et~al.}(2021)Shields, Stevens, Li, Parasram, Damani, Martinez~Alvardo, Janey, Adams, and Doyle]{Shields2021}
B.~J. Shields, J.~Stevens, J.~Li, M.~Parasram, F.~Damani, J.~I. Martinez~Alvardo, J.~M. Janey, R.~P. Adams and A.~G. Doyle, \emph{Nature}, 2021, \textbf{590}, 89--96\relax
\mciteBstWouldAddEndPuncttrue
\mciteSetBstMidEndSepPunct{\mcitedefaultmidpunct}
{\mcitedefaultendpunct}{\mcitedefaultseppunct}\relax
\EndOfBibitem
\bibitem[Angello \emph{et~al.}(2022)Angello, Rathore, Beker, Wołos, Jira, Roszak, Wu, Schroeder, Aspuru-Guzik, Grzybowski, and Burke]{Angello2022}
N.~H. Angello, V.~Rathore, W.~Beker, A.~Wołos, E.~R. Jira, R.~Roszak, T.~C. Wu, C.~M. Schroeder, A.~Aspuru-Guzik, B.~A. Grzybowski and M.~D. Burke, \emph{Science}, 2022, \textbf{378}, 399--405\relax
\mciteBstWouldAddEndPuncttrue
\mciteSetBstMidEndSepPunct{\mcitedefaultmidpunct}
{\mcitedefaultendpunct}{\mcitedefaultseppunct}\relax
\EndOfBibitem
\bibitem[Shi \emph{et~al.}(2023)Shi, Lookman, and Xue]{Shi2023}
B.~Shi, T.~Lookman and D.~Xue, \emph{Materials Genome Engineering Advances}, 2023, \textbf{1}, e14\relax
\mciteBstWouldAddEndPuncttrue
\mciteSetBstMidEndSepPunct{\mcitedefaultmidpunct}
{\mcitedefaultendpunct}{\mcitedefaultseppunct}\relax
\EndOfBibitem
\bibitem[Snoek \emph{et~al.}(2012)Snoek, Larochelle, and Adams]{Snoek2012}
J.~Snoek, H.~Larochelle and R.~P. Adams, Advances in Neural Information Processing Systems, 2012\relax
\mciteBstWouldAddEndPuncttrue
\mciteSetBstMidEndSepPunct{\mcitedefaultmidpunct}
{\mcitedefaultendpunct}{\mcitedefaultseppunct}\relax
\EndOfBibitem
\bibitem[Frazier(2018)]{Frazier2018}
P.~I. Frazier, \emph{A Tutorial on Bayesian Optimization}, 2018, \url{https://arxiv.org/abs/1807.02811}\relax
\mciteBstWouldAddEndPuncttrue
\mciteSetBstMidEndSepPunct{\mcitedefaultmidpunct}
{\mcitedefaultendpunct}{\mcitedefaultseppunct}\relax
\EndOfBibitem
\bibitem[Garnett(2023)]{Garnett2023}
R.~Garnett, \emph{Bayesian Optimization}, Cambridge University Press, 2023\relax
\mciteBstWouldAddEndPuncttrue
\mciteSetBstMidEndSepPunct{\mcitedefaultmidpunct}
{\mcitedefaultendpunct}{\mcitedefaultseppunct}\relax
\EndOfBibitem
\bibitem[Tom \emph{et~al.}(2024)Tom, Schmid, Baird, Cao, Darvish, Hao, Lo, Pablo-García, Rajaonson, Skreta, Yoshikawa, Corapi, Akkoc, Strieth-Kalthoff, Seifrid, and Aspuru-Guzik]{Tom2024}
G.~Tom, S.~P. Schmid, S.~G. Baird, Y.~Cao, K.~Darvish, H.~Hao, S.~Lo, S.~Pablo-García, E.~M. Rajaonson, M.~Skreta, N.~Yoshikawa, S.~Corapi, G.~D. Akkoc, F.~Strieth-Kalthoff, M.~Seifrid and A.~Aspuru-Guzik, \emph{Chem. Rev.}, 2024, \textbf{124}, 9633--9732\relax
\mciteBstWouldAddEndPuncttrue
\mciteSetBstMidEndSepPunct{\mcitedefaultmidpunct}
{\mcitedefaultendpunct}{\mcitedefaultseppunct}\relax
\EndOfBibitem
\bibitem[Strieth-Kalthoff \emph{et~al.}(2024)Strieth-Kalthoff, Hao, Rathore, Derasp, Gaudin, Angello, Seifrid, Trushina, Guy, Liu, Tang, Mamada, Wang, Tsagaantsooj, Lavigne, Pollice, Wu, Hotta, Bodo, Li, Haddadnia, Wołos, Roszak, Ser, Bozal-Ginesta, Hickman, Vestfrid, Aguilar-Granda, Klimareva, Sigerson, Hou, Gahler, Lach, Warzybok, Borodin, Rohrbach, Sanchez-Lengeling, Adachi, Grzybowski, Cronin, Hein, Burke, and Aspuru-Guzik]{StriethKalthoff2024}
F.~Strieth-Kalthoff, H.~Hao, V.~Rathore, J.~Derasp, T.~Gaudin, N.~H. Angello, M.~Seifrid, E.~Trushina, M.~Guy, J.~Liu, X.~Tang, M.~Mamada, W.~Wang, T.~Tsagaantsooj, C.~Lavigne, R.~Pollice, T.~C. Wu, K.~Hotta, L.~Bodo, S.~Li, M.~Haddadnia, A.~Wołos, R.~Roszak, C.~T. Ser, C.~Bozal-Ginesta, R.~J. Hickman, J.~Vestfrid, A.~Aguilar-Granda, E.~L. Klimareva, R.~C. Sigerson, W.~Hou, D.~Gahler, S.~Lach, A.~Warzybok, O.~Borodin, S.~Rohrbach, B.~Sanchez-Lengeling, C.~Adachi, B.~A. Grzybowski, L.~Cronin, J.~E. Hein, M.~D. Burke and A.~Aspuru-Guzik, \emph{Science}, 2024, \textbf{384}, eadk9227\relax
\mciteBstWouldAddEndPuncttrue
\mciteSetBstMidEndSepPunct{\mcitedefaultmidpunct}
{\mcitedefaultendpunct}{\mcitedefaultseppunct}\relax
\EndOfBibitem
\bibitem[Clayton \emph{et~al.}(2019)Clayton, Manson, Taylor, Chamberlain, Taylor, Clemens, and Bourne]{Clayton2019}
A.~D. Clayton, J.~A. Manson, C.~J. Taylor, T.~W. Chamberlain, B.~A. Taylor, G.~Clemens and R.~A. Bourne, \emph{React. Chem. Eng.}, 2019, \textbf{4}, 1545--1554\relax
\mciteBstWouldAddEndPuncttrue
\mciteSetBstMidEndSepPunct{\mcitedefaultmidpunct}
{\mcitedefaultendpunct}{\mcitedefaultseppunct}\relax
\EndOfBibitem
\bibitem[Torres \emph{et~al.}(2022)Torres, Lau, Anchuri, Stevens, Tabora, Li, Borovika, Adams, and Doyle]{Torres2022}
J.~A.~G. Torres, S.~H. Lau, P.~Anchuri, J.~M. Stevens, J.~E. Tabora, J.~Li, A.~Borovika, R.~P. Adams and A.~G. Doyle, \emph{J. Am. Chem. Soc.}, 2022, \textbf{144}, 19999--20007\relax
\mciteBstWouldAddEndPuncttrue
\mciteSetBstMidEndSepPunct{\mcitedefaultmidpunct}
{\mcitedefaultendpunct}{\mcitedefaultseppunct}\relax
\EndOfBibitem
\bibitem[Taylor \emph{et~al.}(2023)Taylor, Pomberger, Felton, Grainger, Barecka, Chamberlain, Bourne, Johnson, and Lapkin]{Taylor2023}
C.~J. Taylor, A.~Pomberger, K.~C. Felton, R.~Grainger, M.~Barecka, T.~W. Chamberlain, R.~A. Bourne, C.~N. Johnson and A.~A. Lapkin, \emph{Chem. Rev.}, 2023, \textbf{123}, 3089--3126\relax
\mciteBstWouldAddEndPuncttrue
\mciteSetBstMidEndSepPunct{\mcitedefaultmidpunct}
{\mcitedefaultendpunct}{\mcitedefaultseppunct}\relax
\EndOfBibitem
\bibitem[Waltz(1967)]{Waltz1967}
F.~Waltz, \emph{IEEE Transactions on Automatic Control}, 1967, \textbf{12}, 179--180\relax
\mciteBstWouldAddEndPuncttrue
\mciteSetBstMidEndSepPunct{\mcitedefaultmidpunct}
{\mcitedefaultendpunct}{\mcitedefaultseppunct}\relax
\EndOfBibitem
\bibitem[Stadler(1988)]{Stadler1988}
W.~Stadler, \emph{Multicriteria Optimization in Engineering and in the Sciences}, Springer Nature, 1988\relax
\mciteBstWouldAddEndPuncttrue
\mciteSetBstMidEndSepPunct{\mcitedefaultmidpunct}
{\mcitedefaultendpunct}{\mcitedefaultseppunct}\relax
\EndOfBibitem
\bibitem[Rentmeesters \emph{et~al.}(1996)Rentmeesters, Tsai, and Lin]{Rentmeesters1996}
M.~Rentmeesters, W.~Tsai and K.-J. Lin, Proceedings of the 2nd IEEE International Conference on Engineering of Complex Computer Systems, 1996, pp. 76--79\relax
\mciteBstWouldAddEndPuncttrue
\mciteSetBstMidEndSepPunct{\mcitedefaultmidpunct}
{\mcitedefaultendpunct}{\mcitedefaultseppunct}\relax
\EndOfBibitem
\bibitem[Deb(2011)]{Deb2011}
K.~Deb, in \emph{Multi-objective Optimisation Using Evolutionary Algorithms: An Introduction}, ed. L.~Wang, A.~H.~C. Ng and K.~Deb, Springer London, London, 2011, pp. 3--34\relax
\mciteBstWouldAddEndPuncttrue
\mciteSetBstMidEndSepPunct{\mcitedefaultmidpunct}
{\mcitedefaultendpunct}{\mcitedefaultseppunct}\relax
\EndOfBibitem
\bibitem[Emmerich \emph{et~al.}(2006)Emmerich, Giannakoglou, and Naujoks]{Emmerich2006}
M.~Emmerich, K.~Giannakoglou and B.~Naujoks, IEEE Transactions on Evolutionary Computation, 2006, pp. 421--439\relax
\mciteBstWouldAddEndPuncttrue
\mciteSetBstMidEndSepPunct{\mcitedefaultmidpunct}
{\mcitedefaultendpunct}{\mcitedefaultseppunct}\relax
\EndOfBibitem
\bibitem[Daulton \emph{et~al.}(2020)Daulton, Balandat, and Bakshy]{Daulton2020}
S.~Daulton, M.~Balandat and E.~Bakshy, Proceedings of the 34th International Conference on Neural Information Processing Systems, 2020\relax
\mciteBstWouldAddEndPuncttrue
\mciteSetBstMidEndSepPunct{\mcitedefaultmidpunct}
{\mcitedefaultendpunct}{\mcitedefaultseppunct}\relax
\EndOfBibitem
\bibitem[Chugh(2020)]{Chugh2020}
T.~Chugh, 2020 IEEE Congress on Evolutionary Computation, 2020, pp. 1--8\relax
\mciteBstWouldAddEndPuncttrue
\mciteSetBstMidEndSepPunct{\mcitedefaultmidpunct}
{\mcitedefaultendpunct}{\mcitedefaultseppunct}\relax
\EndOfBibitem
\bibitem[Klarner \emph{et~al.}(2024)Klarner, Rudner, Morris, Deane, and Teh]{Klarner2024}
L.~Klarner, T.~G.~J. Rudner, G.~M. Morris, C.~Deane and Y.~W. Teh, Proceedings of the 41th International Conference on Machine Learning, 2024\relax
\mciteBstWouldAddEndPuncttrue
\mciteSetBstMidEndSepPunct{\mcitedefaultmidpunct}
{\mcitedefaultendpunct}{\mcitedefaultseppunct}\relax
\EndOfBibitem
\bibitem[Kristiadi \emph{et~al.}(2024)Kristiadi, Strieth-Kalthoff, Skreta, Poupart, Aspuru-Guzik, and Pleiss]{Kristiadi2024}
A.~Kristiadi, F.~Strieth-Kalthoff, M.~Skreta, P.~Poupart, A.~Aspuru-Guzik and G.~Pleiss, Proceedings of the 41th International Conference on Machine Learning, 2024\relax
\mciteBstWouldAddEndPuncttrue
\mciteSetBstMidEndSepPunct{\mcitedefaultmidpunct}
{\mcitedefaultendpunct}{\mcitedefaultseppunct}\relax
\EndOfBibitem
\bibitem[Astudillo and Frazier(2019)]{Astudillo2019}
R.~Astudillo and P.~Frazier, Proceedings of the 36th International Conference on Machine Learning, 2019, pp. 354--363\relax
\mciteBstWouldAddEndPuncttrue
\mciteSetBstMidEndSepPunct{\mcitedefaultmidpunct}
{\mcitedefaultendpunct}{\mcitedefaultseppunct}\relax
\EndOfBibitem
\bibitem[Häse \emph{et~al.}(2021)Häse, Aldeghi, Hickman, Roch, Christensen, Liles, Hein, and Aspuru-Guzik]{Haese2021}
F.~Häse, M.~Aldeghi, R.~J. Hickman, L.~M. Roch, M.~Christensen, E.~Liles, J.~E. Hein and A.~Aspuru-Guzik, \emph{Mach. Learn. Sci. Technol.}, 2021, \textbf{2}, 035021\relax
\mciteBstWouldAddEndPuncttrue
\mciteSetBstMidEndSepPunct{\mcitedefaultmidpunct}
{\mcitedefaultendpunct}{\mcitedefaultseppunct}\relax
\EndOfBibitem
\bibitem[Balandat \emph{et~al.}(2020)Balandat, Karrer, Jiang, Daulton, Letham, Wilson, and Bakshy]{Balandat2020}
M.~Balandat, B.~Karrer, D.~Jiang, S.~Daulton, B.~Letham, A.~G. Wilson and E.~Bakshy, Advances in Neural Information Processing Systems, 2020, pp. 21524--21538\relax
\mciteBstWouldAddEndPuncttrue
\mciteSetBstMidEndSepPunct{\mcitedefaultmidpunct}
{\mcitedefaultendpunct}{\mcitedefaultseppunct}\relax
\EndOfBibitem
\bibitem[Daulton \emph{et~al.}(2022)Daulton, Wan, Eriksson, Balandat, Osborne, and Bakshy]{Daulton2022}
S.~Daulton, X.~Wan, D.~Eriksson, M.~Balandat, M.~A. Osborne and E.~Bakshy, Advances in Neural Information Processing Systems, 2022, pp. 12760 -- 12774\relax
\mciteBstWouldAddEndPuncttrue
\mciteSetBstMidEndSepPunct{\mcitedefaultmidpunct}
{\mcitedefaultendpunct}{\mcitedefaultseppunct}\relax
\EndOfBibitem
\bibitem[Walker \emph{et~al.}(2017)Walker, Bannock, Nightingale, and deMello]{Walker2017}
B.~E. Walker, J.~H. Bannock, A.~M. Nightingale and J.~C. deMello, \emph{React. Chem. Eng.}, 2017, \textbf{2}, 785--798\relax
\mciteBstWouldAddEndPuncttrue
\mciteSetBstMidEndSepPunct{\mcitedefaultmidpunct}
{\mcitedefaultendpunct}{\mcitedefaultseppunct}\relax
\EndOfBibitem
\bibitem[Emmerich \emph{et~al.}(2011)Emmerich, Deutz, and Klinkenberg]{Emmerich2011}
M.~T.~M. Emmerich, A.~H. Deutz and J.~W. Klinkenberg, 2011 IEEE Congress of Evolutionary Computation (CEC), 2011, pp. 2147--2154\relax
\mciteBstWouldAddEndPuncttrue
\mciteSetBstMidEndSepPunct{\mcitedefaultmidpunct}
{\mcitedefaultendpunct}{\mcitedefaultseppunct}\relax
\EndOfBibitem
\bibitem[Häse \emph{et~al.}(2018)Häse, Roch, and Aspuru-Guzik]{Haese2018}
F.~Häse, L.~M. Roch and A.~Aspuru-Guzik, \emph{Chem. Sci.}, 2018, \textbf{9}, 7642--7655\relax
\mciteBstWouldAddEndPuncttrue
\mciteSetBstMidEndSepPunct{\mcitedefaultmidpunct}
{\mcitedefaultendpunct}{\mcitedefaultseppunct}\relax
\EndOfBibitem
\bibitem[Mekki-Berrada \emph{et~al.}(2021)Mekki-Berrada, Ren, Huang, Wong, Zheng, Xie, Tian, Jayavelu, Mahfoud, Bash, Hippalgaonkar, Khan, Buonassisi, Li, and Wang]{mekki-berrada2021nanoparticle}
F.~Mekki-Berrada, Z.~Ren, T.~Huang, W.~K. Wong, F.~Zheng, J.~Xie, I.~P.~S. Tian, S.~Jayavelu, Z.~Mahfoud, D.~Bash, K.~Hippalgaonkar, S.~Khan, T.~Buonassisi, Q.~Li and X.~Wang, \emph{npj Comput. Mater.}, 2021, \textbf{7}, 55\relax
\mciteBstWouldAddEndPuncttrue
\mciteSetBstMidEndSepPunct{\mcitedefaultmidpunct}
{\mcitedefaultendpunct}{\mcitedefaultseppunct}\relax
\EndOfBibitem
\bibitem[Schweidtmann \emph{et~al.}(2018)Schweidtmann, Clayton, Holmes, Badford, Bourne, and Lapkin]{Schweidtmann2018}
A.~M. Schweidtmann, A.~D. Clayton, N.~Holmes, E.~Badford, R.~A. Bourne and A.~A. Lapkin, \emph{Chem. Eng. J.}, 2018, \textbf{352}, 277--282\relax
\mciteBstWouldAddEndPuncttrue
\mciteSetBstMidEndSepPunct{\mcitedefaultmidpunct}
{\mcitedefaultendpunct}{\mcitedefaultseppunct}\relax
\EndOfBibitem
\bibitem[git()]{github}
\url{https:github.com/fsk_lab/botier}\relax
\mciteBstWouldAddEndPuncttrue
\mciteSetBstMidEndSepPunct{\mcitedefaultmidpunct}
{\mcitedefaultendpunct}{\mcitedefaultseppunct}\relax
\EndOfBibitem
\bibitem[Pedregosa \emph{et~al.}(2011)Pedregosa, Varoquaux, Gramfort, Michel, Thirion, Grisel, Blondel, Prettenhofer, Weiss, Dubourg, Vanderplas, Passos, Cournapeau, Brucher, Perrot, and Duchesnay]{scikit-learn}
F.~Pedregosa, G.~Varoquaux, A.~Gramfort, V.~Michel, B.~Thirion, O.~Grisel, M.~Blondel, P.~Prettenhofer, R.~Weiss, V.~Dubourg, J.~Vanderplas, A.~Passos, D.~Cournapeau, M.~Brucher, M.~Perrot and E.~Duchesnay, \emph{J. Mach. Learn. Res.}, 2011, \textbf{12}, 2825--2830\relax
\mciteBstWouldAddEndPuncttrue
\mciteSetBstMidEndSepPunct{\mcitedefaultmidpunct}
{\mcitedefaultendpunct}{\mcitedefaultseppunct}\relax
\EndOfBibitem
\end{mcitethebibliography}
\bibliographystyle{rsc}

\appendix
\begin{si}
\begin{center} 
\bf Supplementary Materials for
\end{center}

\maketitle

\begin{authors}
Mohammad Haddadnia,$^{a, b}$ Leonie Grashoff,$^c$ and Felix Strieth-Kalthoff$^{~c, d,\ast}$
\end{authors}

\begin{affiliations}
\item[$^a$]{Harvard University, Department of Biological Chemistry \& Molecular Pharmacology, Boston (MA), United States.}
\item[$^b$]{Harvard University, Dana-Farber Cancer Institute, Department of Cancer Biology, Boston (MA), United States.}
\item[$^c$]{University of Wuppertal, School of Mathematics and Natural Sciences, Wuppertal, Germany.}
\item[$^d$]{University of Wuppertal, Interdisciplinary Center for Machine Learning and Data Analytics, Wuppertal, Germany.}
\item[$^{\ast}$]{E-Mail: strieth-kalthoff@uni-wuppertal.de}
\end{affiliations}

\startcontents[sections]
\printcontents[sections]{l}{1}{\setcounter{tocdepth}{2}}

\newpage

\section{Implementation Details}
\label{sec:si_implementation}

\subsection{Formulation and implementation of \textit{BoTier}}
\label{sec:si_formulation}

As discussed in the main text (eq. 2), for a scenario in which each objective $\psi_1, ..., \psi_N$ is maximized, we formulate the hierarchical composite objective $\Xi$ as

\begin{equation}
    \Xi = \sum_{i=1}^N \left( \min(\psi_i, t_i) \cdot \prod_{j=1}^{i-1} H(\psi_j - t_j) \right)
\label{eq:si_botier}
\end{equation}

The objectives are sorted by their hierarchy, i.e. $\psi_1$ is the most important objective, $\psi_2$ is the second objective etc. $t_i$ are the corresponding satisfaction thresholds for each objective. The formulation of \textit{BoTier} ensures that, only if an objective $psi_i$ satisfies the respective criterion (i.e., if $\psi_i > t_i$), the subordinate objective $\psi_{i+1}$ is considered. 

$H$ refers to the Heaviside step function:

\begin{equation}
    H(x) = \begin{cases}
    0, & \text{ if } x < 0,\\
    1, & \text{ otherwise}
    \end{cases}
  \label{eq:heaviside}
\end{equation}

To ensure robust optimizability, $\Xi$ should ideally be smooth and differentiable. For these purposes, we use the following analogs for the Heaviside step function\cite{Haese2018} and the $\max$ function, respectively

\begin{equation}
    H(x) \approx \frac{1}{1 + e^{-kx}}
\end{equation}

\begin{equation}
    \max(x_1, x_2) \approx \frac{x_1 \cdot e^{-kx_1} + x_2 \cdot e^{-kx_2}}{e^{-kx_1} + e^{-kx_2}}
\end{equation}

where $k$ is a user-defined "smoothness" parameter. Benchmark experiments for different values of $k$ are provided in Fig.~\ref{fig:ana_botier_posterior_k1_k10_k100} and Fig.~\ref{fig:mod_botier_posterior_k1_k10_k100}.

\subsection{Background: Monte-Carlo Integration and Composite Objectives}
\label{sec:si_montecarlo}

When attempting to optimize a black-box function $f(x)$, BO usually determines the optimal next action to take (i.e., the optimal next $x$ to evaluate) by optimizing an acquisition function $\alpha(x)$.\cite{Garnett2023} This notion originates from the framework of Bayesian decision theory, which identifies the optimal action as the one that results in the maximized expected utility $u$ given all previous observations $\mathcal{D}$.\cite{Garnett2023}

\begin{equation}
    x_{\text{next}} \in \argmax{x \in \mathcal{X}}{\EX{u(x, f(x), \theta)}{\mathcal{D}}}
\label{eq:bayes_decision}
\end{equation}

The utility function $u$ defines the user's preference over what would be the ideal next observation, which can be a function of the inputs $x$, the value of the function $f(x)$, and additional parameters $\theta$. For simple utility functions (such as the improvement $u = \max{(f(x) -f^\ast, 0)}$ and Gaussian posterior distributions $p(f(x) \mid \mathcal{D})$, this expectation value can be calculated analytically – e.g., for the improvement utility, the corresponding acquisition function is the well-known \textit{Expected Improvement}. For utility functions where the calculation of this expectation value is not analytically tractable, Monte-Carlo sampling allows for approximating the expectation value in eq. \ref{eq:bayes_decision} by drawing $K$ samples $\xi_k(x)$ from the posterior belief $p(f(x) \mid \mathcal{D})$.

\begin{equation}
    \EX{u(x, f(x), \theta)}{\mathcal{D}} \approx \frac{1}{K} \sum_{k=1}^K u(x, \xi_k(x), \theta) 
\label{eq:mc_acqf}
\end{equation}

This strategy has been popularized by Balandat \etal with the development of \textit{BoTorch} \cite{Balandat2020}, which allows for differentiable (and therefore, readily optimizable) acquisition function calculation through Monte-Carlo sampling. 

The framework of approximating the expectation value in eq. \ref{eq:bayes_decision} through Monte-Carlo methods can be extended to a scenario in which the overall optimization goal is decoupled from the observable functions over which we have a posterior belief. 
Given a set of observable functions $\left\{ f_i(x) \right\}_{i=1}^N$, we introduce the optimization goal $g$ as $g\big(x, f_1(x), ..., f_N(x) \big)$.
Using the Monte-Carlo approximation discussed above, any arbitrary $g$ can be chosen, and the expectation value can be approximated as 

\begin{equation}
  \begin{split}
          &\EX{u\Big(x, g\big(x, f_1(x), ..., f_N(x)\big), \theta \Big)}{\mathcal{D}} \\
          &\approx \frac{1}{K} \sum_{k=1}^K u\Big(x, g\big(x, \xi_{k1}(x), ..., \xi_{kn}(x)\big), \theta \Big) 
  \end{split}
\label{eq:mc_arbitrary_goal}
\end{equation}

with posterior samples $\xi_{ki}(x) \sim p(f_i(x) \mid \mathcal{D})$.

In practical BO setting, we can use a user-defined scalarization function for $g$ as a composite objective.
For all relevant experiment outputs, we can train individual or joint probabilistic surrogate models, which allow us for flexibly encoding any prior knowledge. 
After drawing samples from the posterior distribution(s) of these models, the optimization goal (i.e. the composite score) can be calculated per sample from both input- and output-derived objectives. 
Computing the utility function, and averaging over all samples yields the desired expectation value.

\newpage

\section{Benchmarks on Analytical Multiobjective Problems}
\label{sec:si_benchmarks}

Benchmarks were performed on four multiobjective surfaces, as implemented in the \textit{BoTorch} library: BNH, DH4, DTLZ5 and ZDT1.\citep{Balandat2020} Satisfaction thresholds for each objective were selected to yield challenging, but solvable optimization problems (see section \ref{sec:si_surfaces_analytical} for further details). 

In addition, to simulate optimization problems with preferences over experiment inputs, the above mentioned optimization problems were augmented by fully input-dependent objectives (see section \ref{sec:si_surfaces_analytical} for further details). The respective optimization problems are referred to as BNH*, DH4*, DTLZ5* and ZDT1*, respectively. 

\subsection{Analytical Multiobjective Problems}
\label{sec:si_surfaces_analytical}

Table \ref{tab:analytical_functions} provides an overview of all surfaces and the defined optimization objectives for multi-objective optimization. Figures \ref{fig:bnh_a}–\ref{fig:zdt1_b} show the density of objective values across the surface, along with the defined thresholds. 

\begin{table}[h]
\caption{Overview of the analytical test functions from BoTorch that were used for benchmarking the described optimization strategies. For each function, the first row describes the analytical function with only output-dependent objectives, and the second row includes an additional input-dependent objective. Visualizations of the objective ranges and inter-objective correlations are provided in the following figures.}
\centering
\begin{tabular}{cccccc}
\toprule
Function               & Number of Inputs    & Number of Outputs  & Objective \#1 & Objective \#2            & Objective \#3 \\ \midrule \midrule 
BNH                    & \multirow{2}{*}{2}  & \multirow{2}{*}{2} & $y_0 > -60.0$ & $y_1 > -11.0$            &               \\
BNH*                   &                     &                    & $y_0 > -60.0$ & $x_0 - x_1 > 2.0$ & $y_1 > -15.0$ \\ \midrule
DH4                    & \multirow{2}{*}{6}  & \multirow{2}{*}{2} & $y_0 > -0.15$ & $y_1 > -15.0$            &               \\
DH4*                   &                     &                    & $y_0 > -0.15$ & $x_1 > 0.6$              & $y_1 > -15.0$ \\ \midrule
DTLZ5                  & \multirow{2}{*}{4}  & \multirow{2}{*}{2} & $y_0 > -0.5$  & $y_1 > -0.95$            &               \\
DTLZ5*                 &                     &                    & $y_0 > -0.5$  & $x_2 + x_3 < 1.0$        & $y_1 > -0.95$ \\ \midrule
ZDT1                   & \multirow{2}{*}{10} & \multirow{2}{*}{2} & $y_0 > -0.18$ & $y_1 > -2.5$             &               \\
ZDT1*                  &                     &                    & $y_0 > -0.18$ & $x_1 + x_5 < 0.5$        & $y_1 > -2.5$ \\
\bottomrule
\end{tabular}
\label{tab:analytical_functions}
\vspace{2cm}
\end{table}

\begin{figure}[H]    
 \centering
 \includegraphics[width=18cm]{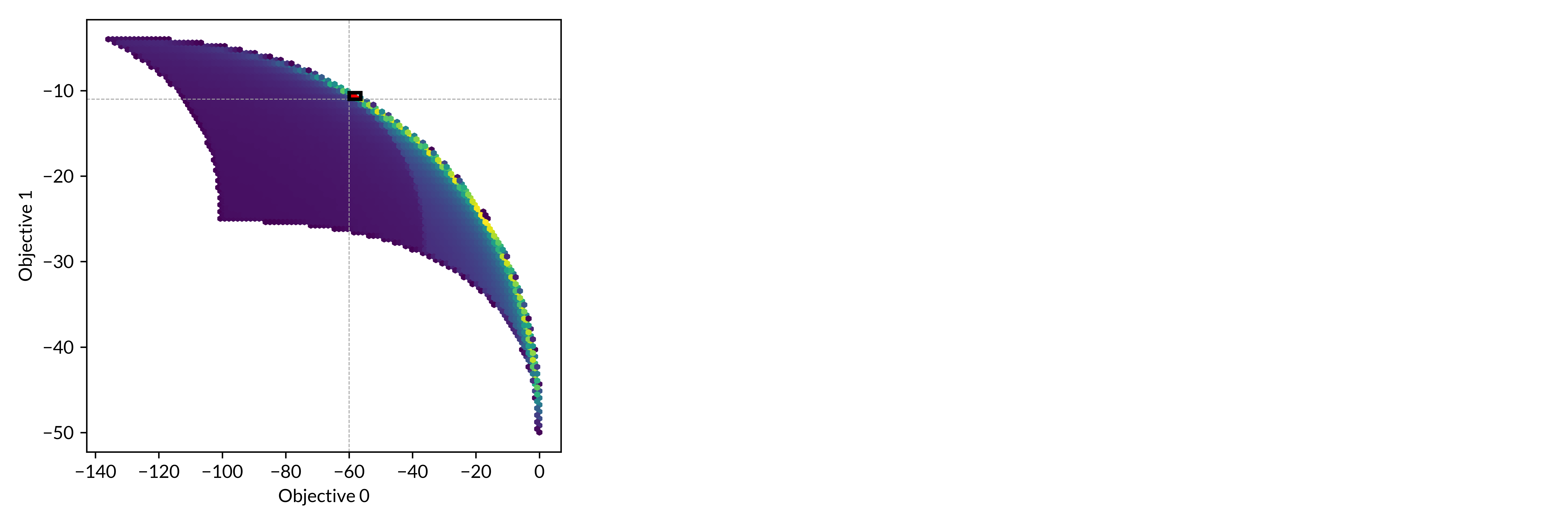}
 \caption{Objective density plots of the analytical BNH surface, as described in Tab. \ref{tab:analytical_functions} and implemented in BoTorch.\cite{Balandat2020} The function was evaluated on a grid of $5\cdot10^6$ points drawn from a Sobol sampler. Data points that satisfy all objectives are shown in red. }
 \label{fig:bnh_a}
\end{figure}

\begin{figure}[H]    
 \centering
 \includegraphics[width=18cm]{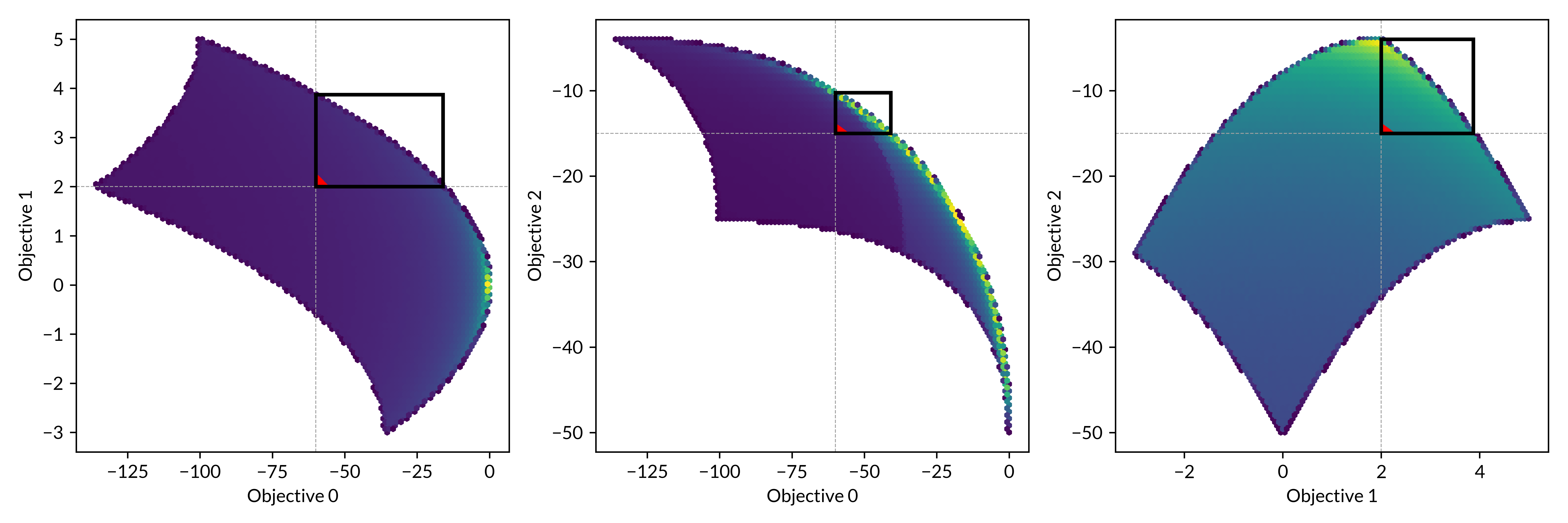}
 \caption{Objective density plots of the modified analytical BNH surface with an additional input-dependent objective, as described in Tab. \ref{tab:analytical_functions}. The function was evaluated on a grid of $5\cdot10^6$ points drawn from a Sobol sampler. Data points that satisfy all objectives are shown in red.}
 \label{fig:bnh_b}
\end{figure}

\begin{figure}[H]    
 \centering
 \includegraphics[width=18cm]{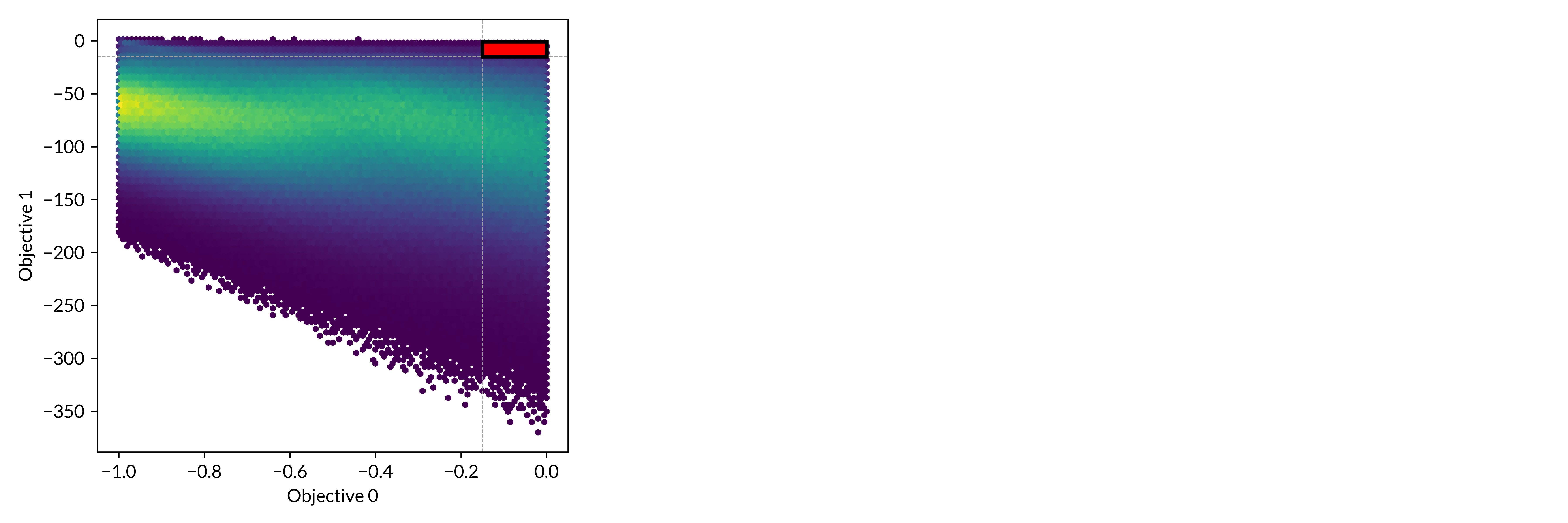}
 \caption{Objective density plots of the analytical DH4 surface, as described in Tab. \ref{tab:analytical_functions} and implemented in BoTorch.\cite{Balandat2020} The function was evaluated on a grid of $5\cdot10^6$ points drawn from a Sobol sampler. Data points that satisfy all objectives are shown in red.}
 \label{fig:dh4_a}
\end{figure}

\begin{figure}[H]    
 \centering
 \includegraphics[width=18cm]{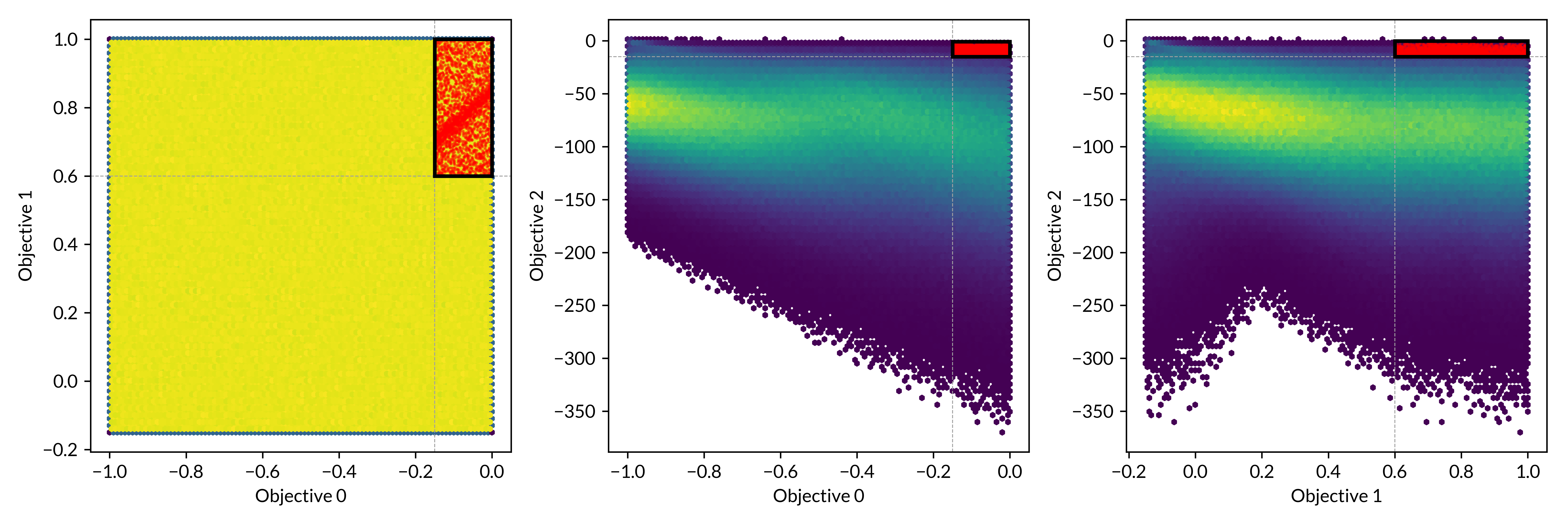}
 \caption{Objective density plots of the modified analytical DH4 surface with an additional input-dependent objective, as described in Tab. \ref{tab:analytical_functions}. The function was evaluated on a grid of $5\cdot10^6$ points drawn from a Sobol sampler. Data points that satisfy all objectives are shown in red.}
 \label{fig:dh4_b}
\end{figure}

\begin{figure}[H]    
 \centering
 \includegraphics[width=18cm]{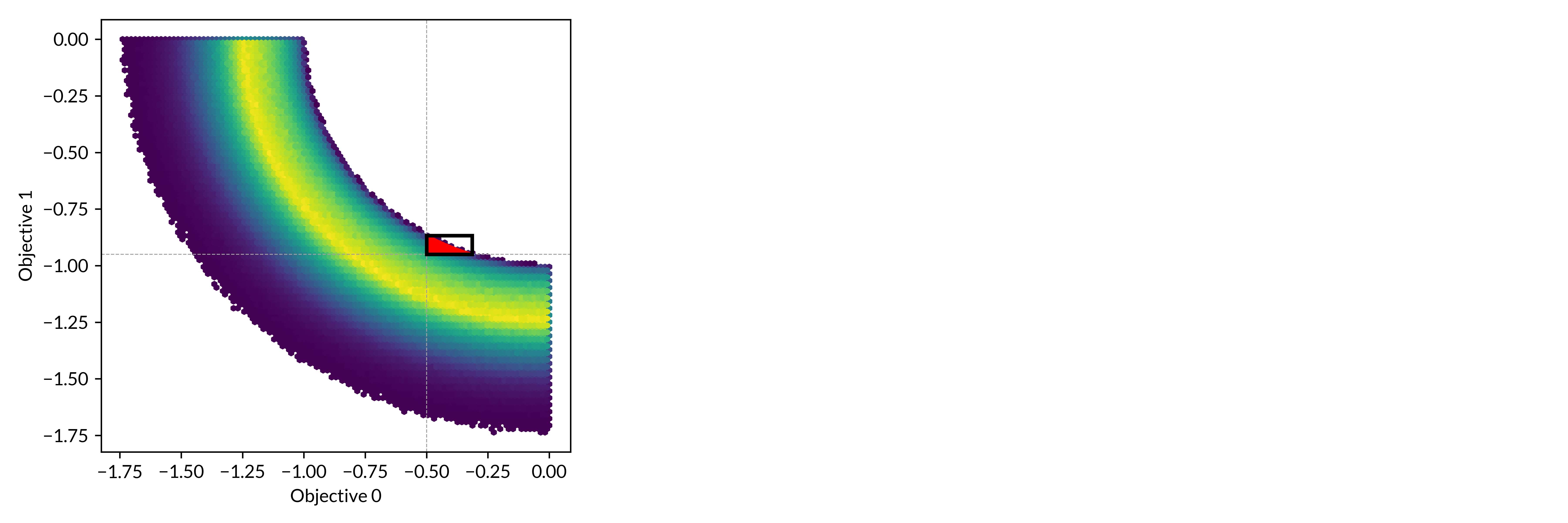}
 \caption{Objective density plots of the analytical DTLZ5 surface, as described in Tab. \ref{tab:analytical_functions} and implemented in BoTorch.\cite{Balandat2020} The function was evaluated on a grid of $5\cdot10^6$ points drawn from a Sobol sampler. Data points that satisfy all objectives are shown in red.}
 \label{fig:dtlz5_a}
\end{figure}

\begin{figure}[H]    
 \centering
 \includegraphics[width=18cm]{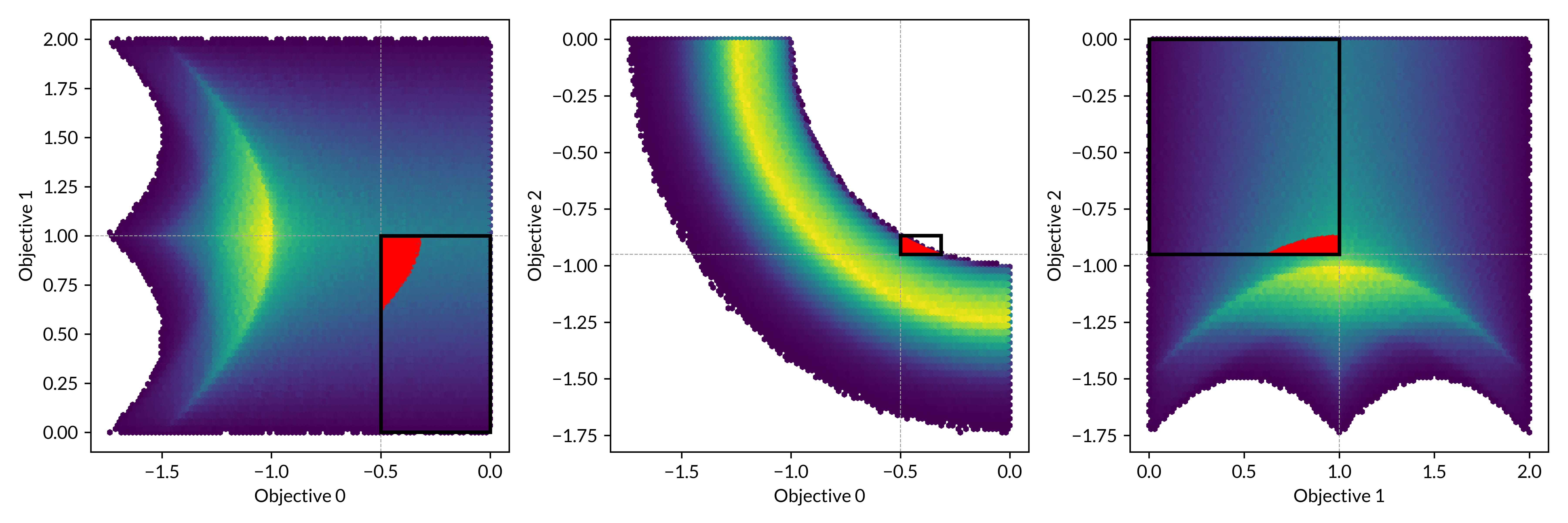}
 \caption{Objective density plots of the modified analytical DTLZ5 surface with an additional input-dependent objective, as described in Tab. \ref{tab:analytical_functions}. The function was evaluated on a grid of $5\cdot10^6$ points drawn from a Sobol sampler. Data points that satisfy all objectives are shown in red.}
 \label{fig:dtlz5_b}
\end{figure}

\begin{figure}[H]    
 \centering
 \includegraphics[width=18cm]{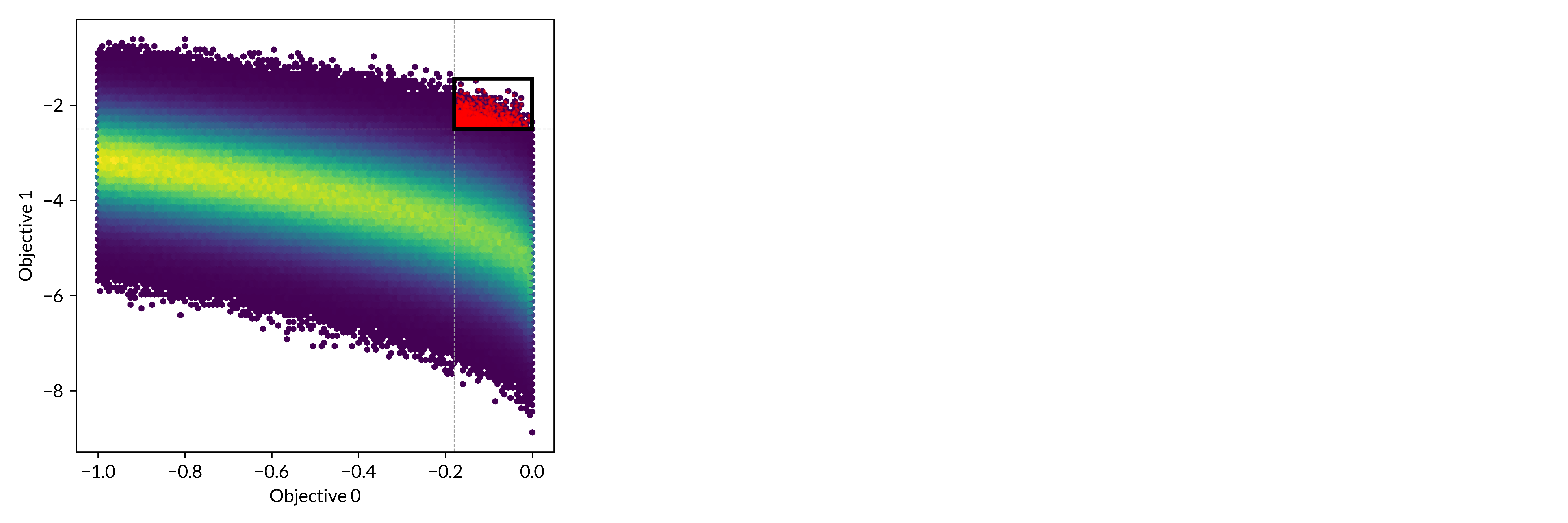}
 \caption{Objective density plots of the analytical ZDT1 surface, as described in Tab. \ref{tab:analytical_functions} and implemented in BoTorch.\cite{Balandat2020} The function was evaluated on a grid of $5\cdot10^6$ points drawn from a Sobol sampler. Data points that satisfy all objectives are shown in red.}
 \label{fig:zdt1_a}
\end{figure}

\begin{figure}[H]    
 \centering
 \includegraphics[width=18cm]{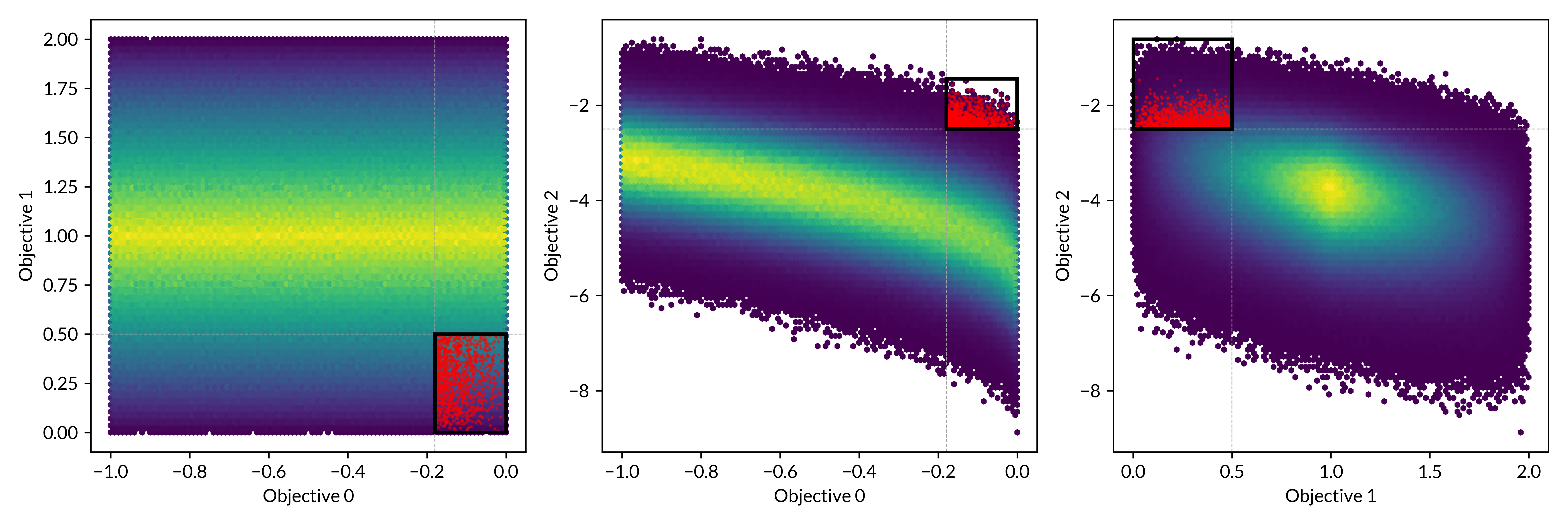}
 \caption{Objective density plots of the modified analytical ZDT1 surface with an additional input-dependent objective, as described in Tab. \ref{tab:analytical_functions}. The function was evaluated on a grid of $5\cdot10^6$ points drawn from a Sobol sampler. Data points that satisfy all objectives are shown in red.}
 \label{fig:zdt1_b}
\end{figure}

\newpage

\subsection{Benchmark Runs on Analytical Surfaces with Solely Output-Dependent Objectives}

All benchmark experiments were performed using the \textit{BoTorch} ecosystem.\citep{Balandat2020} Each optimization campaign starts by randomly drawing a single seed experiment from the respective surface, followed by training a Gaussian Process surrogate model on the obtained data. Using this surrogate, an acquisition function (Expected Improvement, EI or Expected Hypervolume Improvement, EHVI) is optimized over the full input parameter space to obtain a single ($q=1$) recommendation. This recommendation is evaluated on the analytical surface. With the obtained data, the loop is repeated until a budget of 50 experiments is exhausted. Each optimization campaign is repeated over 50 independent runs, and deviations are visualized based on the standard error across these runs. 

As a model-free, non-iterative baseline, the available budget of experimental recommendations is drawn from a Sobol sampler.

All code to reproduce the experiments is provided on the \textit{GitHub} repository.\citep{github}

\label{sec:si_benchmarks_analytical}

\begin{figure}[H]    
 \centering
 \includegraphics[width=18cm]{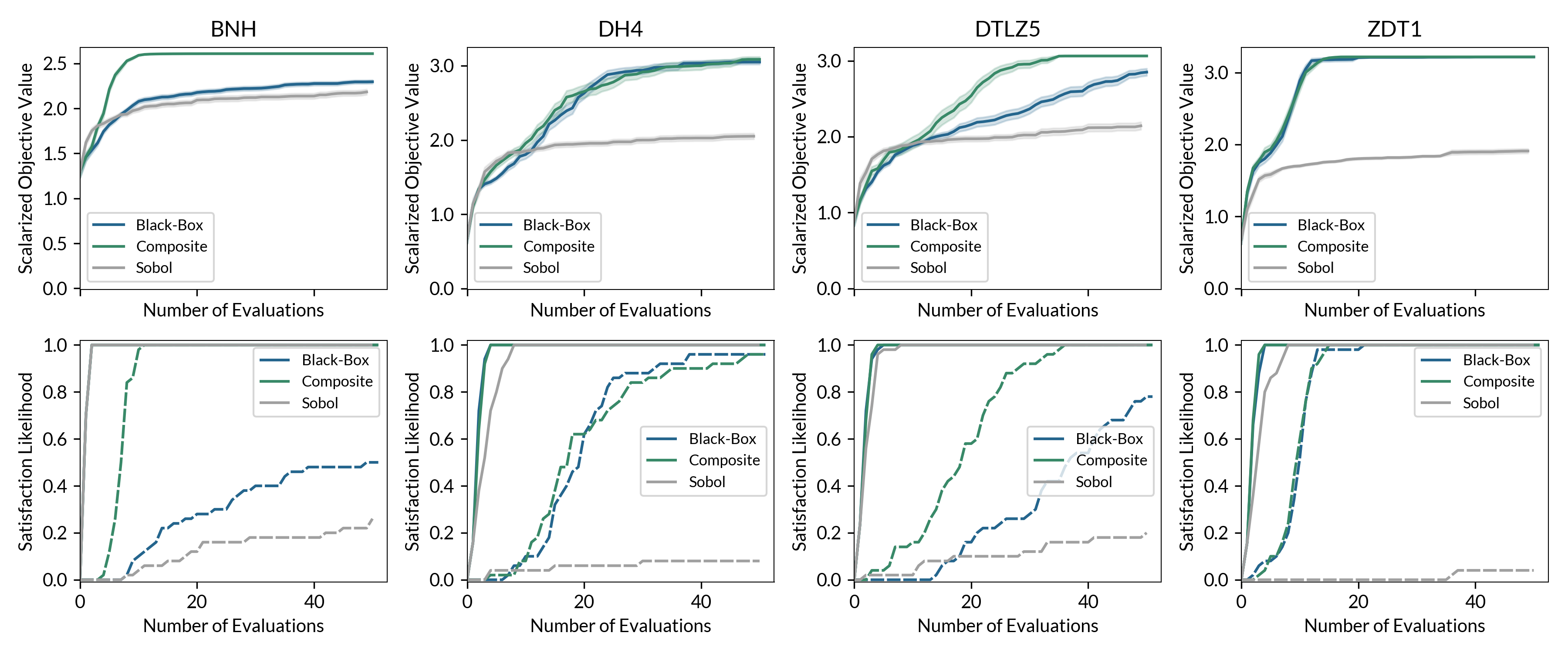}
 \caption{Bayesian Optimization campaigns using \textit{BOTier} as the hierarchical scalarization objective, comparing its use as a composite and a black-box objective. Experiments drawn from a Sobol sampler are included as a non-iterative, model-free baseline. The top row shows the best observed value of $\Xi$ as a function of the number of experiments. The bottom row shows the likelihood of finding a point that satisfies the first (solid line), or the first and the second (dashed line) objective. A single-task Gaussian process was used to model each outcome / objective. Expected Improvement was used as the acquisition function. All statistics were calculated over 50 independent runs, and intervals are plotted based on the standard error.}
 \label{fig:ana_prior-vs-posterior_multi_botier-ei_k100}
\end{figure}

\begin{figure}[H]    
 \centering
 \includegraphics[width=18cm]{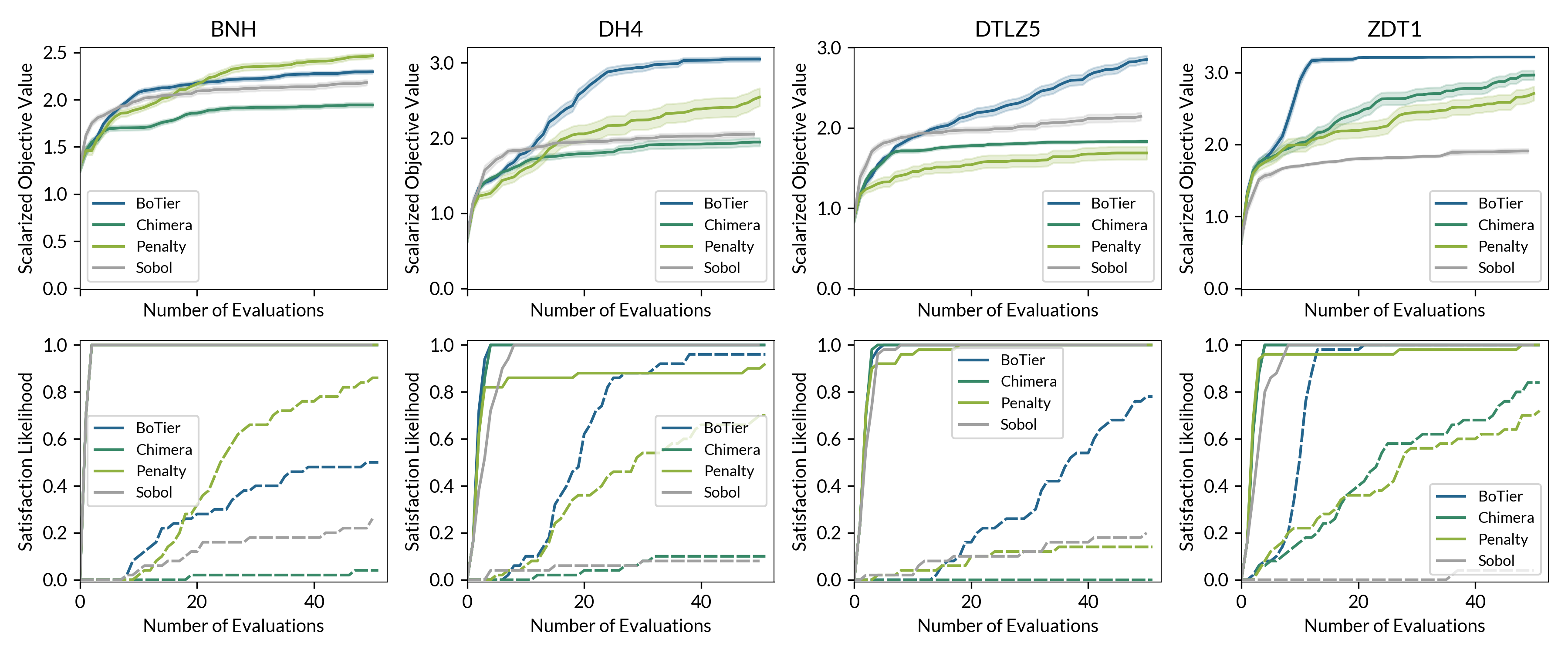}
 \caption{Bayesian Optimization campaigns with black-box objectives, comparing \textit{BOTier}, \textit{Chimera} and a penalty-based scalarization. Experiments drawn from a Sobol sampler are included as a non-iterative, model-free baseline. The top row shows the best observed value of $\Xi$ as a function of the number of experiments. The bottom row shows the likelihood of finding a point that satisfies the first (solid line), or the first and the second (dashed line) objective. A single-task Gaussian process was used to model each outcome / objective. Expected Improvement was used as the acquisition function. All statistics were calculated over 50 independent runs, and intervals are plotted based on the standard error.}
 \label{fig:ana_prior_multi_k100_sampling}
\end{figure}

\begin{figure}[H]    
 \centering
 \includegraphics[width=18cm]{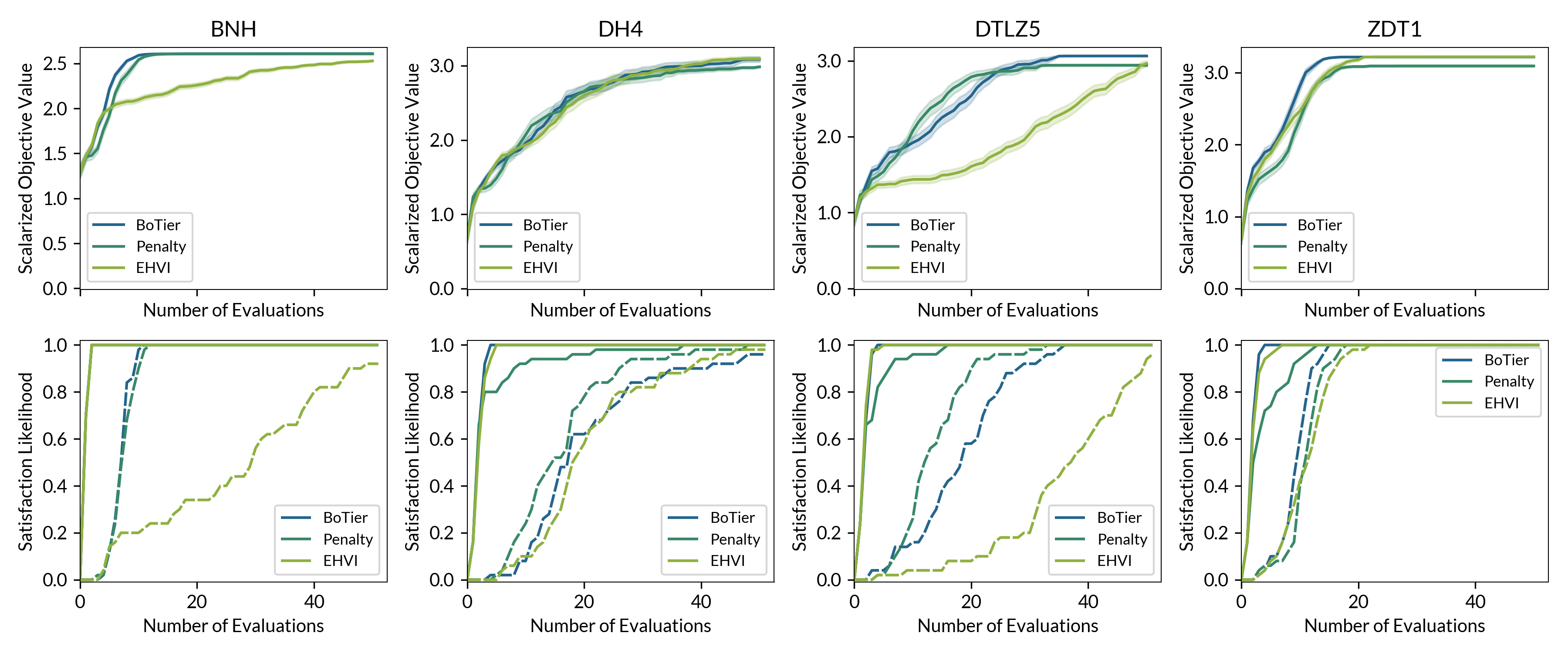}
 \caption{Bayesian Optimization campaigns with black-box objectives, comparing \textit{BOTier}, a penalty-based scalarization, and the Pareto-oriented \textit{Expected Hypervolume Improvement} (EHVI). The top row shows the best observed value of $\Xi$ as a function of the number of experiments. The bottom row shows the likelihood of finding a point that satisfies the first (solid line), or the first and the second (dashed line) objective. A single-task Gaussian process was used to model each outcome / objective. Expected Improvement was used as the acquisition function. All statistics were calculated over 50 independent runs, and intervals are plotted based on the standard error.}
 \label{fig:ana_botier-vs-penalty-vs-ehvi_multi_botier-ei_k100}
\end{figure}

\begin{figure}[H]    
 \centering
 \includegraphics[width=18cm]{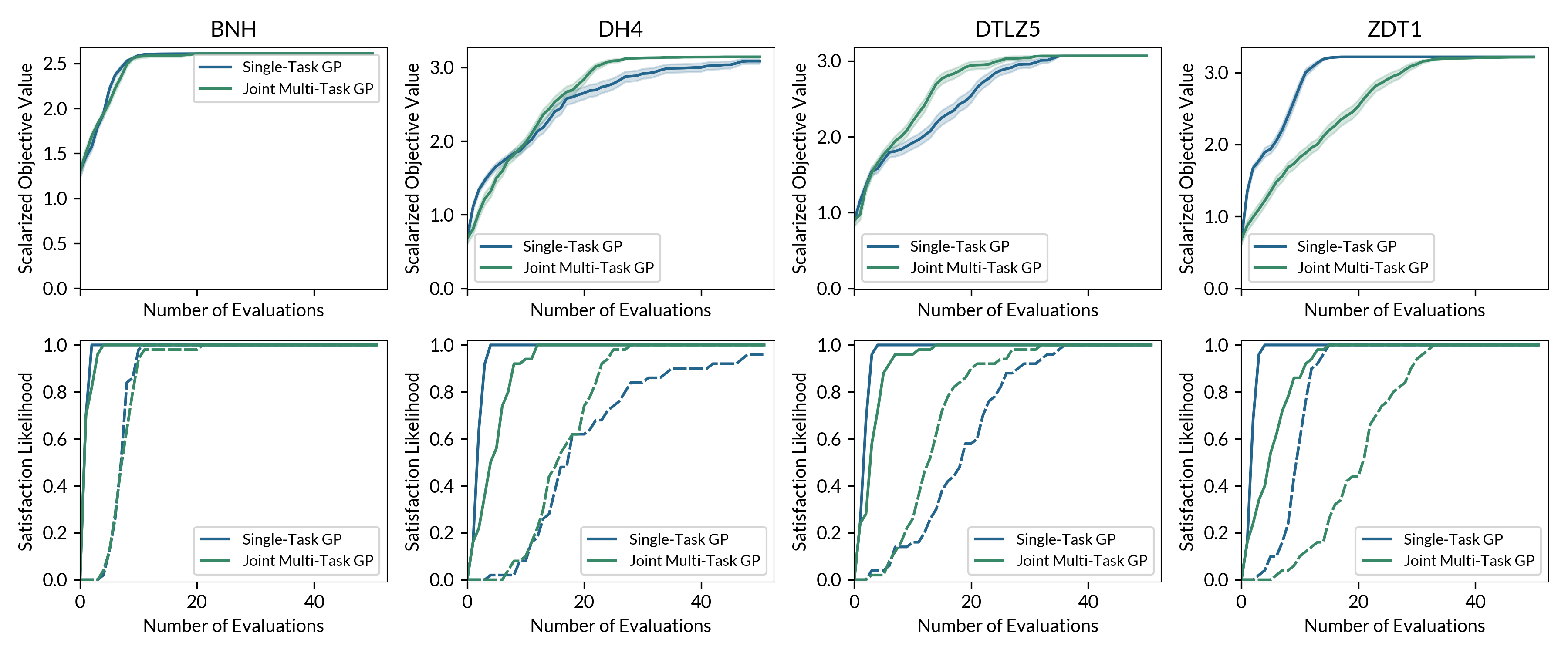}
 \caption{Bayesian Optimization campaigns using \textit{BOTier} as a composite objective, comparing the use of independent Gaussian Process models for each objective ("Single-Task GP) against a joint multi-output Gaussian Process ("Joint Multi-Task GP"). The top row shows the best observed value of $\Xi$ as a function of the number of experiments. The bottom row shows the likelihood of finding a point that satisfies the first (solid line), or the first and the second (dashed line) objective. A single-task Gaussian process was used to model each outcome / objective. Expected Improvement was used as the acquisition function. All statistics were calculated over 50 independent runs, and intervals are plotted based on the standard error.}
\label{fig:joint_surf}
\end{figure}

\begin{figure}[H]    
 \centering
 \includegraphics[width=18cm]{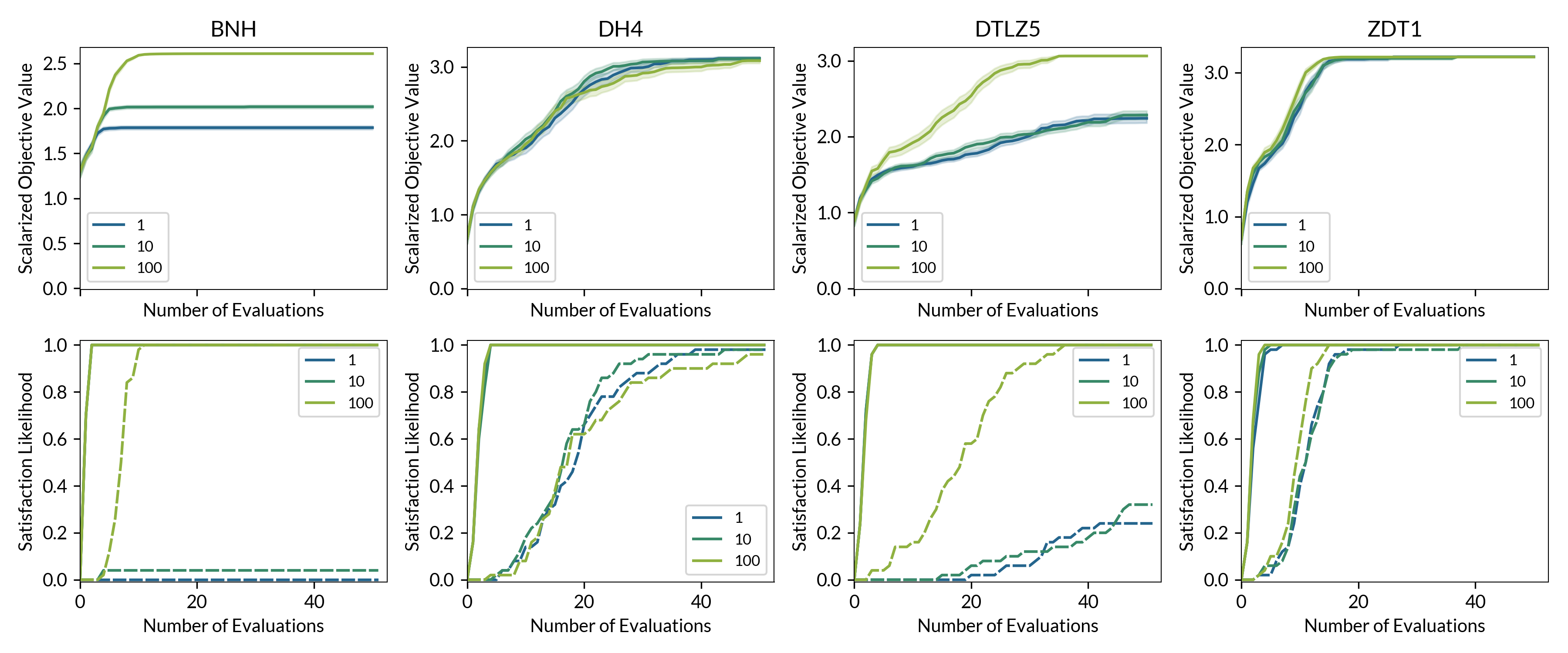}
 \caption{Bayesian Optimization campaigns using \textit{BOTier} as a composite objective, evaluating the impact of the smoothing parameter $k$. The top row shows the best observed value of $\Xi$ as a function of the number of experiments. The bottom row shows the likelihood of finding a point that satisfies the first (solid line), or the first and the second (dashed line) objective. A single-task Gaussian process was used to model each outcome / objective. Expected Improvement was used as the acquisition function. All statistics were calculated over 50 independent runs, and intervals are plotted based on the standard error.}
 \label{fig:ana_botier_posterior_k1_k10_k100}
\end{figure}

\newpage

\subsection{Benchmark Runs on Analytical Surfaces with Output- and Input-Dependent Objectives}
\label{sec:si_benchmarks_analytical_extended}

Benchmark experiments were performed as described in section \ref{sec:si_benchmarks_analytical}. All code to reproduce the experiments is provided on the \textit{GitHub} repository.\citep{github}

\begin{figure}[H]    
 \centering
 \includegraphics[width=18cm]{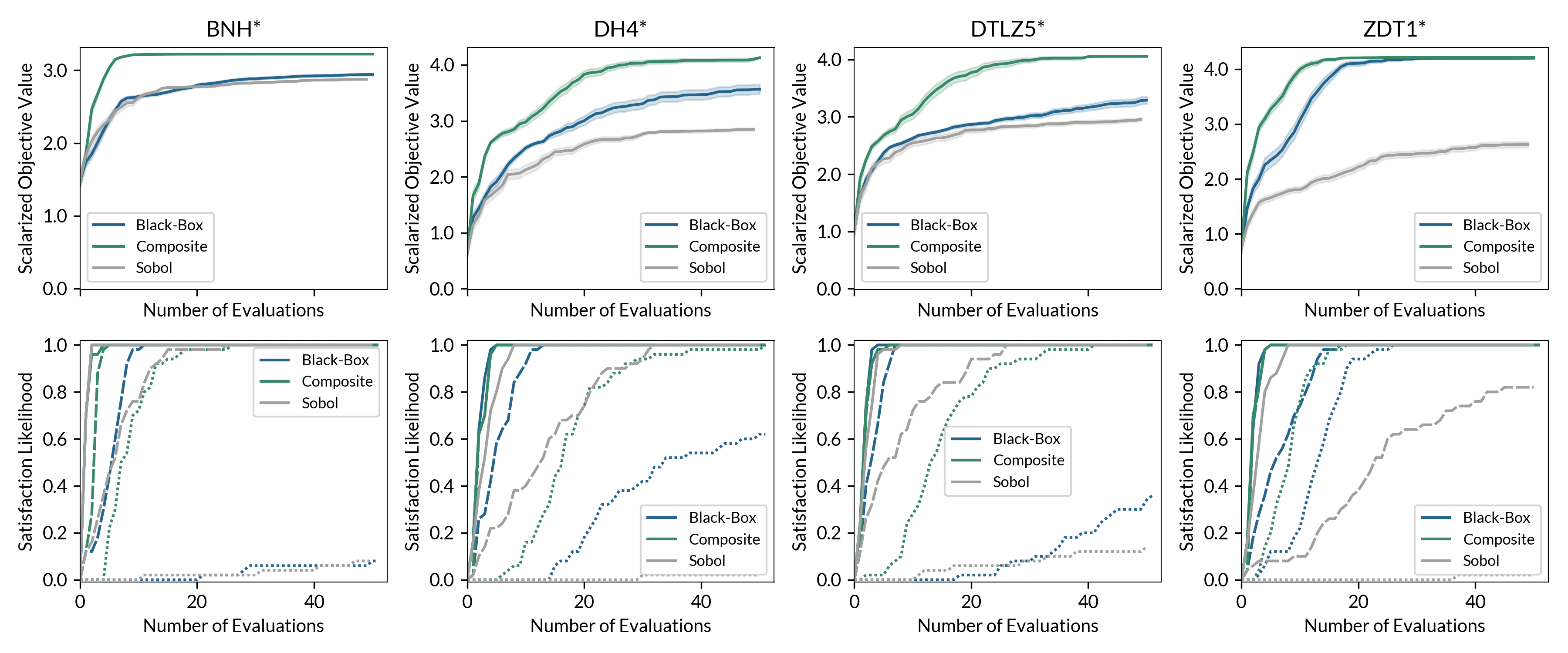}
 \caption{Bayesian Optimization campaigns using \textit{BOTier} as the hierarchical scalarization objective, comparing its use as a composite and a black-box objective. Experiments drawn from a Sobol sampler are included as a non-iterative, model-free baseline. The top row shows the best observed value of $\Xi$ as a function of the number of experiments. The bottom row shows the likelihood of finding a point that satisfies the first (solid line), or the first and the second (dashed line) objective. A single-task Gaussian process was used to model each outcome / objective. Expected Improvement was used as the acquisition function. All statistics were calculated over 50 independent runs, and intervals are plotted based on the standard error.}
 \label{fig:mod_prior-vs-posterior_multi_botier-ei_k100}
\end{figure}

\begin{figure}[H]    
 \centering
 \includegraphics[width=18cm]{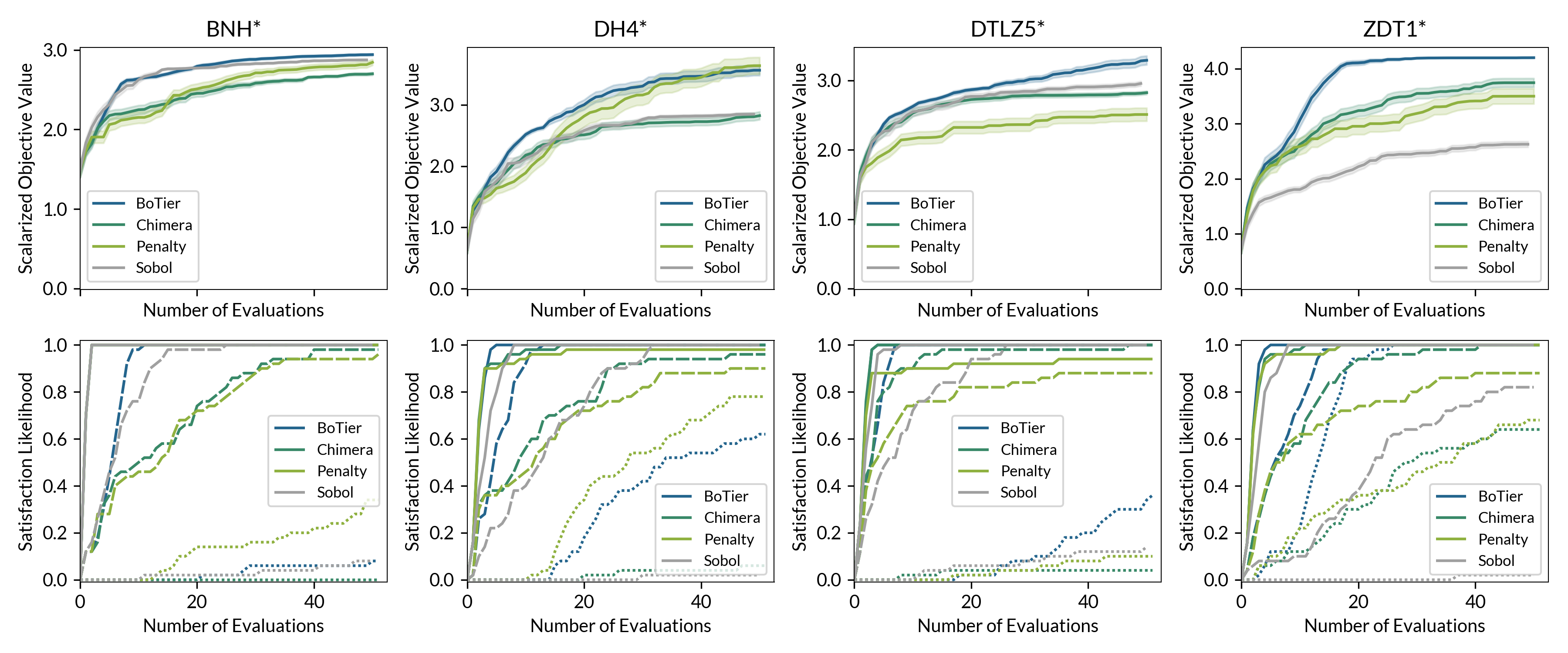}
 \caption{
 Bayesian Optimization campaigns with black-box objectives, comparing \textit{BOTier}, \textit{Chimera} and a penalty-based scalarization. Experiments drawn from a Sobol sampler are included as a non-iterative, model-free baseline. The top row shows the best observed value of $\Xi$ as a function of the number of experiments. The bottom row shows the likelihood of finding a point that satisfies the first (solid line), or the first and the second (dashed line) objective. A single-task Gaussian process was used to model each outcome / objective. Expected Improvement was used as the acquisition function. All statistics were calculated over 50 independent runs, and intervals are plotted based on the standard error.}
 \label{fig:mod_prior_multi_k100_sampling}
\end{figure}

\begin{figure}[H]    
 \centering
 \includegraphics[width=18cm]{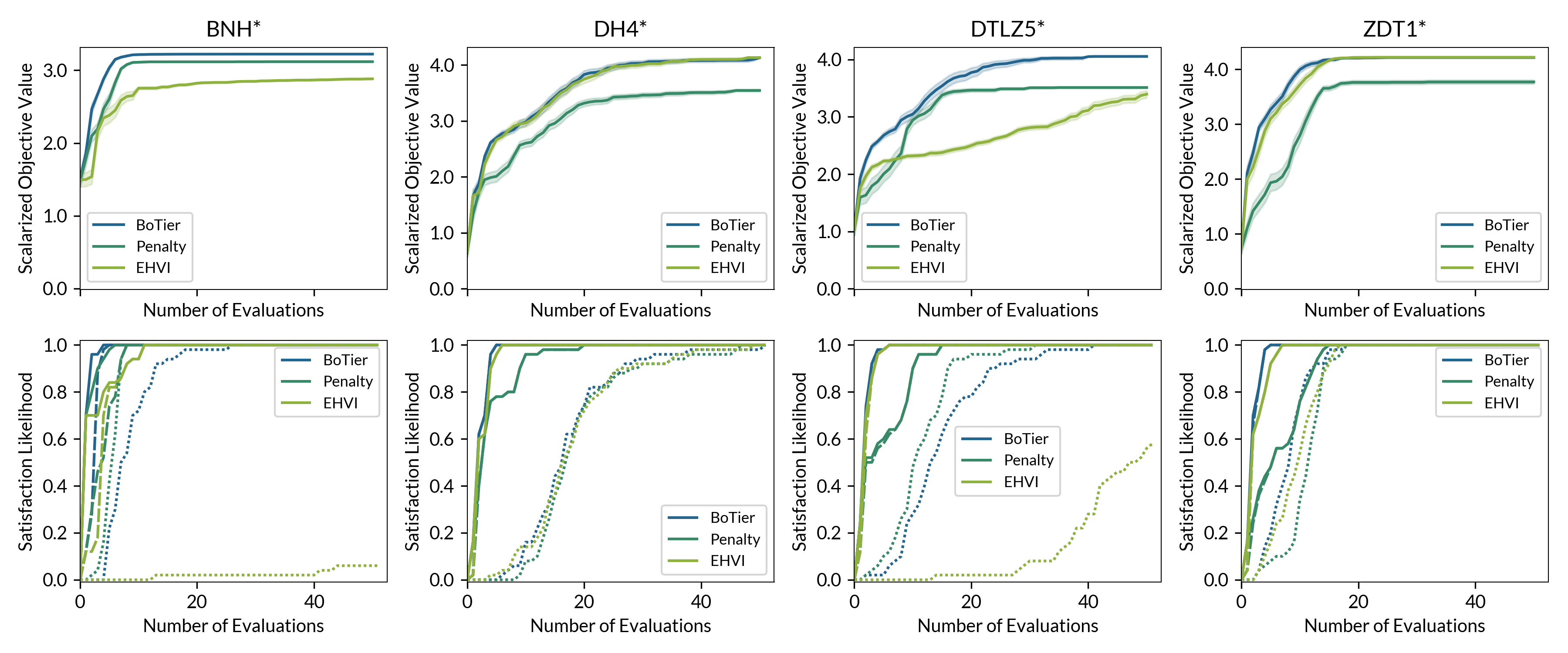}
 \caption{Bayesian Optimization campaigns with black-box objectives, comparing \textit{BOTier}, a penalty-based scalarization, and the Pareto-oriented \textit{Expected Hypervolume Improvement} (EHVI). The top row shows the best observed value of $\Xi$ as a function of the number of experiments. The bottom row shows the likelihood of finding a point that satisfies the first (solid line), or the first and the second (dashed line) objective. A single-task Gaussian process was used to model each outcome / objective. Expected Improvement was used as the acquisition function. All statistics were calculated over 50 independent runs, and intervals are plotted based on the standard error.}
 \label{fig:mod_botier-vs-penalty-vs-ehvi_multi_botier-ei_k100}
\end{figure}

\begin{figure}[H]    
 \centering
 \includegraphics[width=18cm]{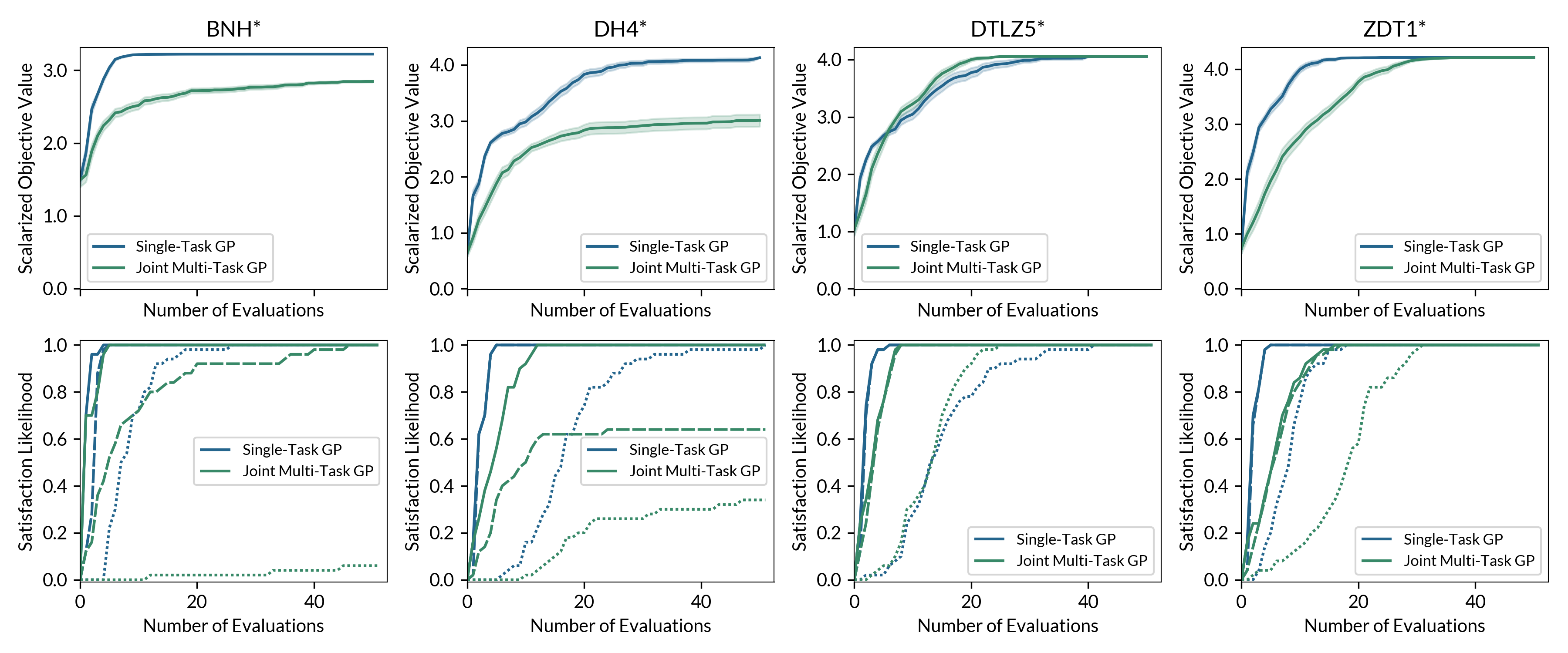}
 \caption{Bayesian Optimization campaigns using \textit{BOTier} as a composite objective, comparing the use of independent Gaussian Process models for each objective ("Single-Task GP) against a joint multi-output Gaussian Process ("Joint Multi-Task GP"). The top row shows the best observed value of $\Xi$ as a function of the number of experiments. The bottom row shows the likelihood of finding a point that satisfies the first (solid line), or the first and the second (dashed line) objective. A single-task Gaussian process was used to model each outcome / objective. Expected Improvement was used as the acquisition function. All statistics were calculated over 50 independent runs, and intervals are plotted based on the standard error.}
\label{fig:joint_mod}
\end{figure}

\begin{figure}[H]    
 \centering
 \includegraphics[width=18cm]{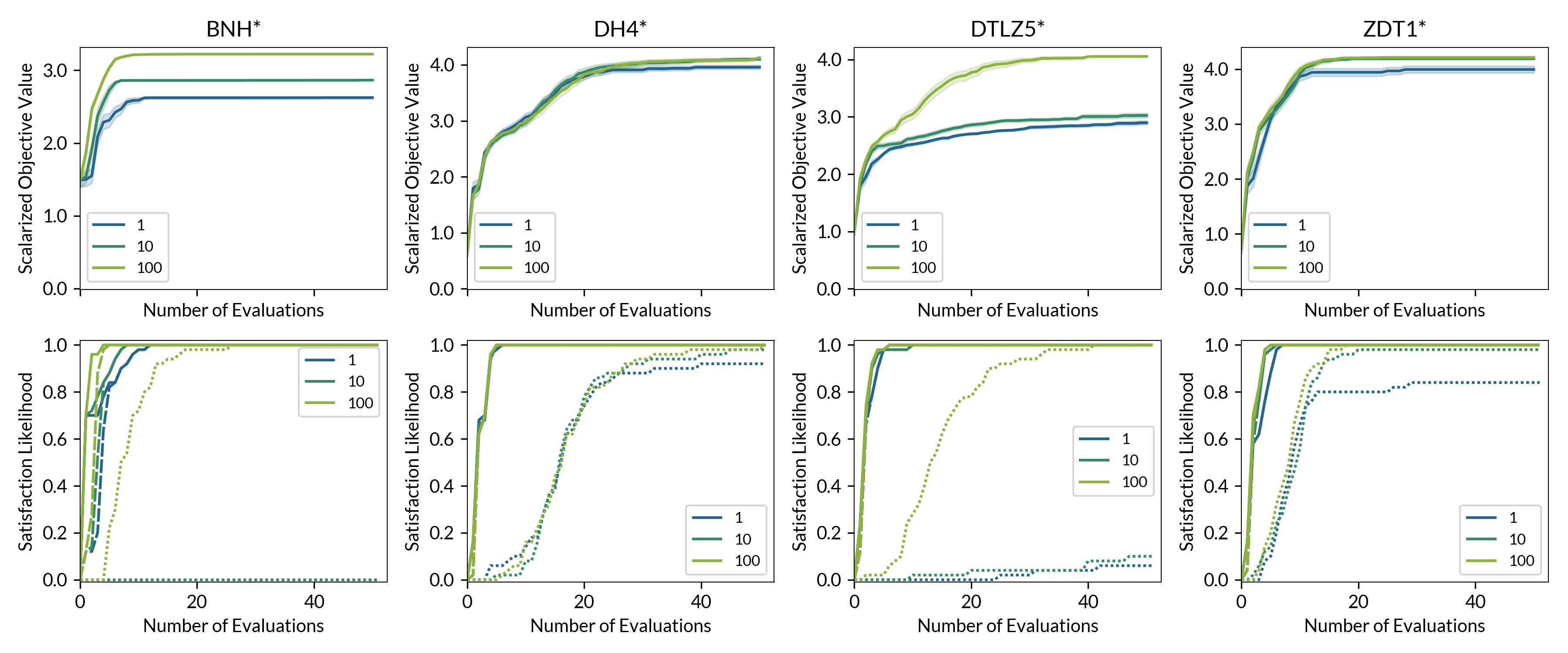}
 \caption{Bayesian Optimization campaigns using \textit{BOTier} as a composite objective, evaluating the impact of the smoothing parameter $k$. The top row shows the best observed value of $\Xi$ as a function of the number of experiments. The bottom row shows the likelihood of finding a point that satisfies the first (solid line), or the first and the second (dashed line) objective. A single-task Gaussian process was used to model each outcome / objective. Expected Improvement was used as the acquisition function. All statistics were calculated over 50 independent runs, and intervals are plotted based on the standard error.}
 \label{fig:mod_botier_posterior_k1_k10_k100}
\end{figure}

\newpage

\section{Benchmarks on Emulated Multiobjective Problems}
\label{sec:si_benchmarks_emulated}

Experiments on real-life problems from chemical reaction optimization were performed following the strategy described by Häse \etal:\citep{Haese2021} Using a series of experimental observations, a regression model is trained to emulate experiments. For the problems analyzed herein, we used a Random Forest regressor or a $k$-Nearest Neighbors Regressor (as implemented in \textit{scikit-learn} \citep{scikit-learn}) as the emulator. Predictive performance was evaluated using 5-fold cross validation, and the best-performing model was selected as the emulator for all following investigations.

Specifically, the following optimization problems were evaluated:
\begin{itemize}
\item{Yield optimization in a heterocyclic Suzuki–Miyaura cross coupling\citep{Haese2021}} 
\item{Yield optimization for the benzylation of $\alpha$-methylbenzylamine in continuous-flow\citep{Schweidtmann2018}} 
\item{Yield optimization for an enzymatic alkoxylation reaction \citep{Haese2021}}
\item{Optimization of the spectral composition of silver nanoparticles\citep{mekki-berrada2021nanoparticle}}
\end{itemize}

For each of these tasks, we use the cumulative cost of materials as a secondary, solely input-dependent objective. The used reagent prices are given in Tab. \ref{tab:si_prices}.

Benchmark experiments were performed as described in section \ref{sec:si_benchmarks_analytical}. The code to reproduce all experiments is provided on our GitHub repository.\citep{github}

\begin{table}[H]
    \caption{Reagent prices in € mol$^{-1}$ (for reactants and reagents) or € L$^{-1}$ (for solvents) used in this study, as obtained from \textit{Sigma-Aldrich} and \textit{ABCR} on July 15, 2024.}
    \centering
    \begin{tabular}{cc}
      \toprule
      Reagent & Price \\ \midrule \midrule
      $\alpha$-Methylbenzylamine & 31 € mol$^{-1}$ \\
      Benzyl bromide & 58 € mol$^{-1}$ \\
      {[1,1$'$-Bis(di-tert-butylphosphino)ferrocen]}dichlorpalladium(II) (Pd(dtbpf)Cl$_2$) & 131,700 € mol$^{-1}$ \\
      Chloroform & 104 € L$^{-1}$ \\
      2,4-Difluoronitrobenzene & 164 € mol$^{-1}$ \\
      Ethanol & 55 € L$^{-1}$ \\
      Morpholine & 21 € mol$^{-1}$ \\
      Potassium phosphate & 20 € mol$^{-1}$ \\
      4-Trifluoromethylphenylboronic acid pinacol ester (CF$_3$-Ph-Bpin) & 940 € mol$^{-1}$ \\
      \bottomrule
    \end{tabular}
    \label{tab:si_prices}
\end{table}

\subsection{Heterocyclic Suzuki–Miyaura Coupling}

\begin{table}[H]
  \caption{Controllable variables for the Heterocyclic Suzuki–Miyaura Coupling}.
  \centering
  \begin{tabularx}{18cm}{cXccc}
  \toprule
  Index & Variable Description & Unit & Lower Bound & Upper Bound \\ \midrule \midrule
  0 & Reaction temperature & $^{\circ}$C & 75.0 & 90.0 \\
  1 & Palladium catalyst loading &  mol-\% & 0.5 & 5.0 \\
  2 & Equivalents of pinacolyl boronate ester coupling partner & - & 1.0 & 1.8 \\
  3 & Equivalents of tripotassium phosphate & - & 1.5 & 3.0 \\
  \bottomrule
  \end{tabularx}
  \label{tab:suzuki_inputs}
\end{table}

\begin{table}[H]
  \caption{Optimization objectives for the Yield Prediction of the heterocyclic Suzuki–Miyaura coupling.}
  \centering
  \begin{tabularx}{18cm}{cccYc}
    \toprule
     & Objective & Direction & Calculation & Threshold \\ \midrule \midrule
    1 & Yield & $\uparrow$ & $y$ & 65.0 \\
    2 & Reaction Cost & $\downarrow$ & $x_1 \cdot 0.01 \cdot p^M_{\text{Pd(dtbpf)Cl2}} + x_2 \cdot p^M_{\text{CF3-Ph-Bpin}} + x_3 \cdot p^M_{\text{K3PO4}}$ & 3500 \\
    3 & Reaction temperature & $\downarrow$ & $x_0$ & 85.0 \\
    \bottomrule
  \end{tabularx}
  \label{tab:suzuki_objectives}
\end{table}

\begin{figure}[H]    
 \centering
 \includegraphics[width=18cm]{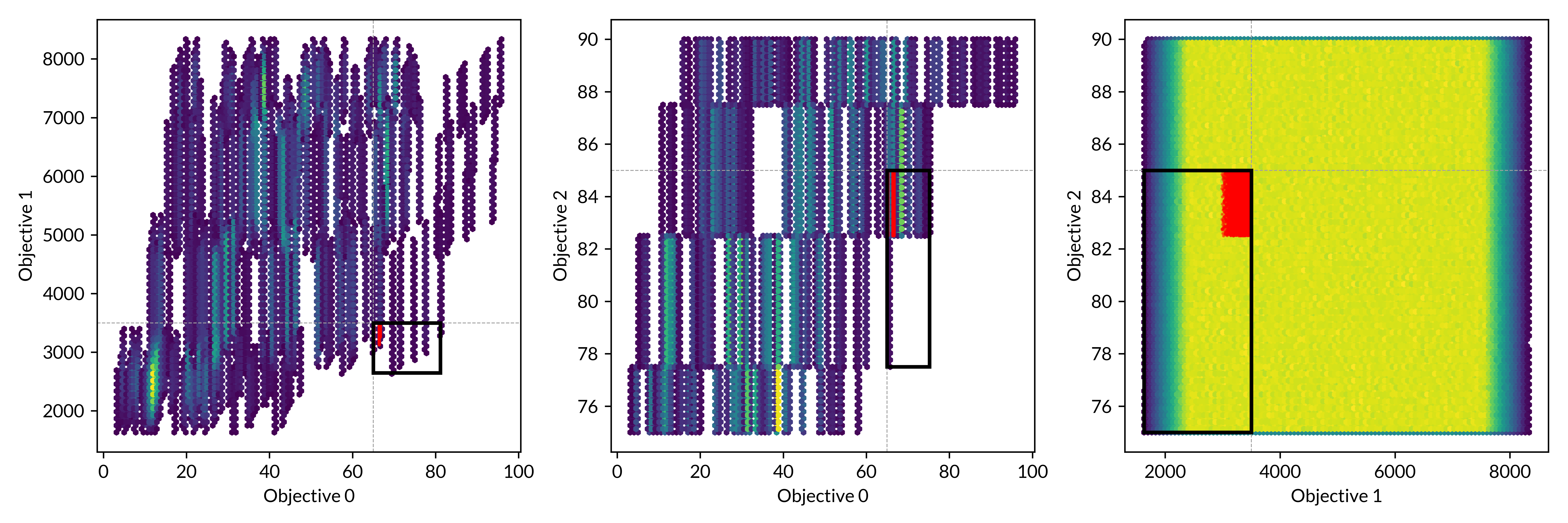}
 \caption{Objective density plots of the Heterocyclic Suzuki–Miyaura Coupling, as described in Tab. \ref{tab:suzuki_inputs} and Tab. \ref{tab:suzuki_objectives}. The function was evaluated on a grid of $5\cdot10^6$ points drawn from a Sobol sampler. Data points that satisfy all objectives are shown in red.}
 \label{fig:suzuki_obj}
\end{figure}

\begin{figure}[H]    
 \centering
 \includegraphics[width=18cm]{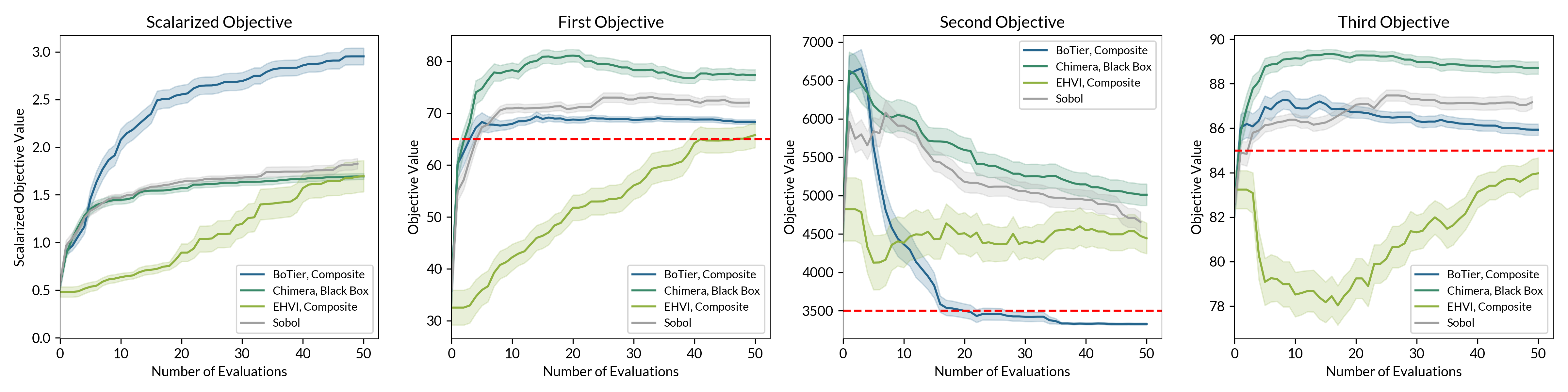}
 \caption{Comparison of different objective functions for multi-objective Bayesian Optimization: \textit{BoTier} (composite objective); \textit{Chimera} (black-box objective); and \textit{Expected Hypervolume Improvement}. All experiments were performed using  single-task Gaussian process model and the Expected Improvement acquisition function, if applicable. Experiments drawn from a Sobol sampler are included as a non-iterative, model-free baseline. All statistics were calculated over 50 independent runs, and intervals are plotted based on the standard error. Panel 1: Trajectories of the best hierarchical objective score obtained after $n$ experiments. Panel 2: Reaction yield at the best conditions found after $n$ experiments. Panel 3: Reagent cost at the best conditions found after $n$ experiments. Panel 4: Reaction temperature at the best conditions found after $n$ experiments.}  
 \label{fig:suzuki_em}
\end{figure}

\newpage

\subsection{Benzylation of $\alpha$-methylbenzylamine}

\begin{table}[H]
  \caption{Controllable variables for the benzylation of $\alpha$-methylbenzylamine with benzyl bromide}.
  \centering
  \begin{tabularx}{18cm}{cXccc}
  \toprule
  Index & Variable Description & Unit & Lower Bound & Upper Bound \\ \midrule \midrule
  0 & Flow rate of $\alpha$-methylbenzylamine solution (0.455 M) & mL min$^{-1}$ & 0.2 & 0.4 \\
  1 & Equivalents of benzyl bromide & - & 1.0 & 5.0 \\
  2 & Ratio of extra solvent : $\alpha$-methylbenzylamine & - & 0.5 & 1.0 \\
  3 & Reaction temperature & $^{\circ}$C & 110 & 150 \\
  \bottomrule
  \end{tabularx}
  \label{tab:benzylation_inputs}
\end{table}

\begin{table}[H]
    \caption{Optimization objectives for the benzylation of $\alpha$-methylbenzylamine with benzyl bromide.}
    \centering
    \begin{tabularx}{18cm}{cccYc}
      \toprule
       & Objective & Direction & Calculation & Threshold \\ \midrule \midrule
      1 & Yield of dibenzylation product & $\uparrow$ & $y$ & 4.0 \\
      2 & Reagent Cost & $\downarrow$ & $x_0 \cdot 0.455 \cdot p^M_{\alpha\text{-Methylbenzylamine}} + x_0 \cdot 0.455 \cdot x_1 \cdot p^M_{\text{Benzyl bromide}}$ $+ x_0 \cdot (1 + x_1 + x_2) \cdot p^V_{\text{Chloroform}}$  & 100 \\
      3 & Temperature & $\downarrow$ & $x_3$ & 115 \\
      \bottomrule 
    \end{tabularx}
    \label{tab:benzylation_objectives}
\end{table}

\begin{figure}[H]    
 \centering
 \includegraphics[width=18cm]{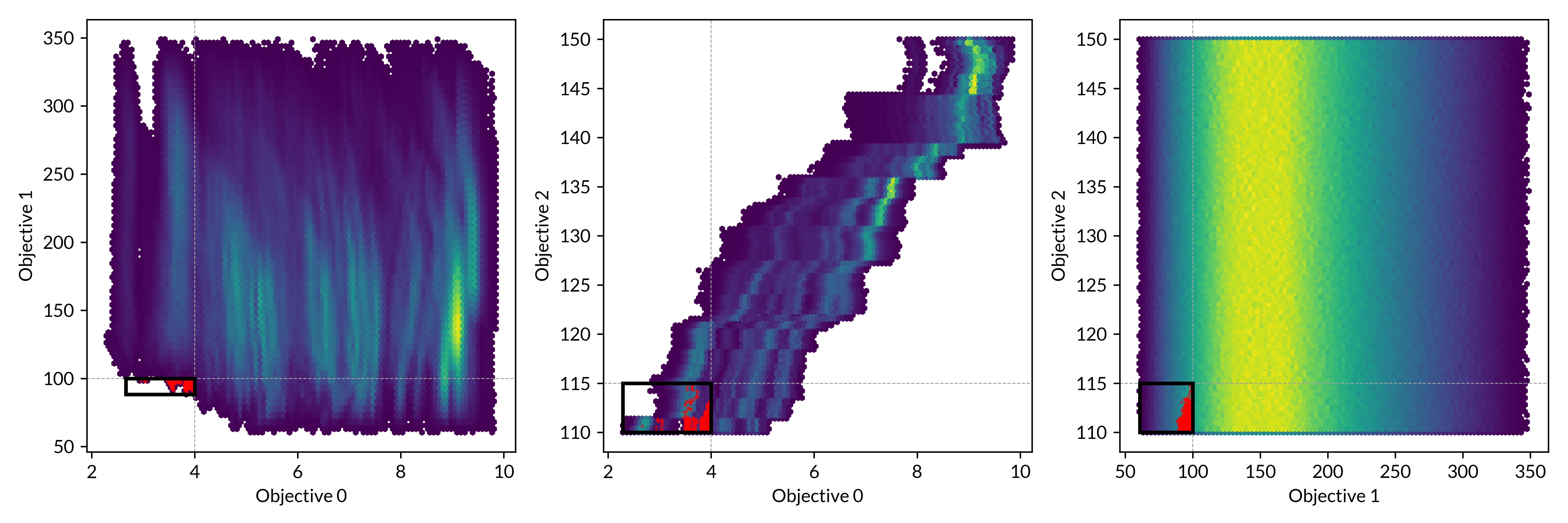}
 \caption{Objective density plots of the benzylation of $\alpha$-methylbenzylamine, as described in Tab. \ref{tab:benzylation_inputs} and Tab. \ref{tab:benzylation_objectives}. The function was evaluated on a grid of $5\cdot10^6$ points drawn from a Sobol sampler. Data points that satisfy all objectives are shown in red.}
 \label{fig:benzylation_obj}
\end{figure}

\begin{figure}[H]    
 \centering
 \includegraphics[width=18cm]{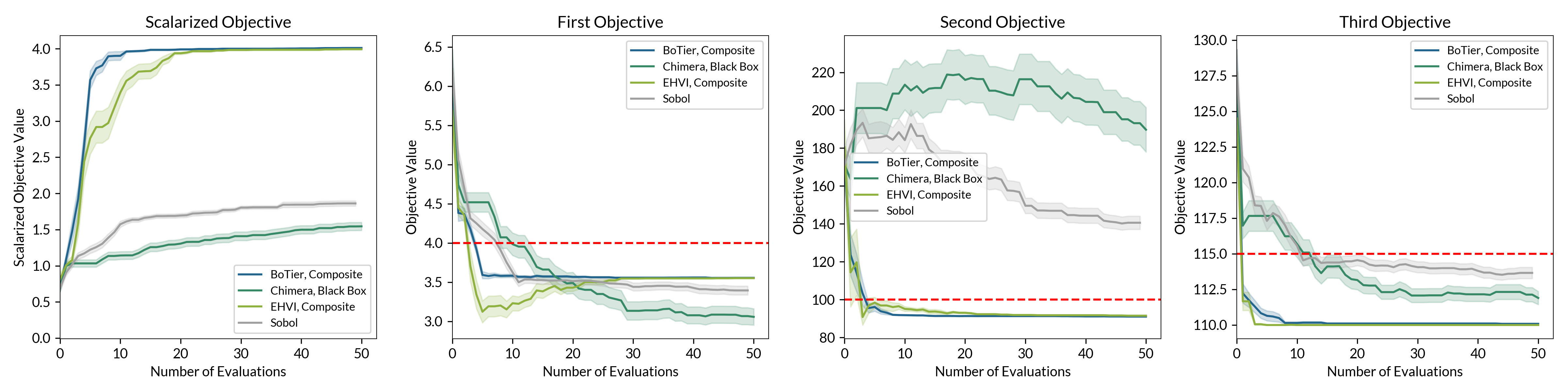}
 \caption{Comparison of different objective functions for multi-objective Bayesian Optimization: \textit{BoTier} (composite objective); \textit{Chimera} (black-box objective); and \textit{Expected Hypervolume Improvement}. All experiments were performed using  single-task Gaussian process model and the Expected Improvement acquisition function, if applicable. Experiments drawn from a Sobol sampler are included as a non-iterative, model-free baseline. All statistics were calculated over 50 independent runs, and intervals are plotted based on the standard error. Panel 1: Trajectories of the best hierarchical objective score obtained after $n$ experiments. Panel 2: Reaction yield at the best conditions found after $n$ experiments. Panel 3: Reagent cost at the best conditions found after $n$ experiments. Panel 4: Reaction temperature at the best conditions found after $n$ experiments.}  
 \label{fig:benz_em}
\end{figure}

\newpage

\subsection{Enzymatic Alkoxylation}

\begin{table}[H]
  \caption{Controllable variables for the Enzymatic Alkoxylation}.
  \begin{tabularx}{18cm}{cXccc}
  \toprule
  Index & Variable Description & Unit & Lower Bound & Upper Bound \\ \midrule \midrule
      0 & Catalase concentration & $\mu$M & 0.05 & 1.0 \\
      1 & Horseradish peroxidase  & $\mu$M & 0.5 & 10.0 \\
      2 & Alcohol oxidase concentration & nM & 2.0 & 8.0  \\
      3 & pH & - & 6.0 & 8.0 \\
    \bottomrule
  \end{tabularx}
  \label{tab:alkoxylation_inputs}
\end{table}

\begin{table}[H]
  \caption{Optimization objectives for the enzymatic alkoxylation yield prediction.}
  \centering
  \begin{tabularx}{18cm}{cccYc}
    \toprule
     & Objective & Direction & Calculation & Threshold \\ \midrule \midrule
        1 & Yield & $\uparrow$ & $y$ & 20.0 \\
        2 & Enzyme Quantities & $\downarrow$ & $x_0 + x_1 + x_2$ & 10.0 \\
    \bottomrule
  \end{tabularx}
  \label{tab:alkoxylation_objectives}
\end{table}

\begin{figure}[H]    
 \centering
 \includegraphics[width=18cm]{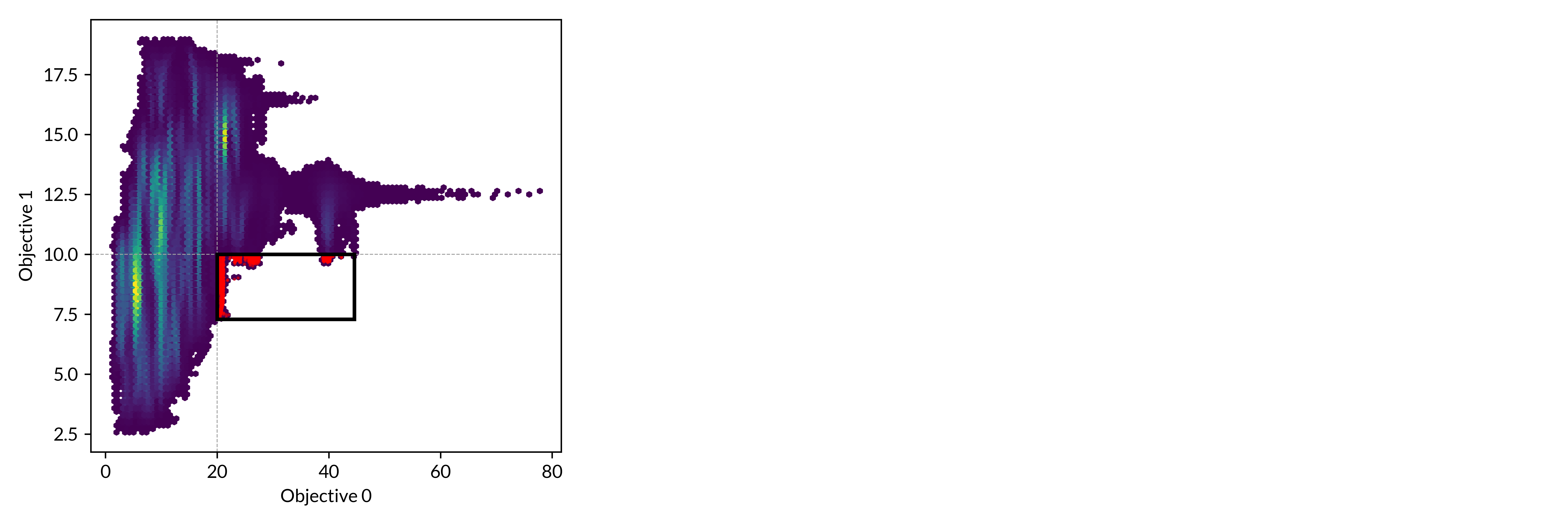}
 \caption{Objective density plots of the Enzymatic Alkoxylation, as described in Tab. \ref{tab:alkoxylation_inputs} and Tab. \ref{tab:alkoxylation_objectives}. The function was evaluated on a grid of $5\cdot10^6$ points drawn from a Sobol sampler. Data points that satisfy all objectives are shown in red.}
 \label{fig:Alkoxylation_obj}
\end{figure}

\begin{figure}[H]    
 \centering
 \includegraphics[width=18cm]{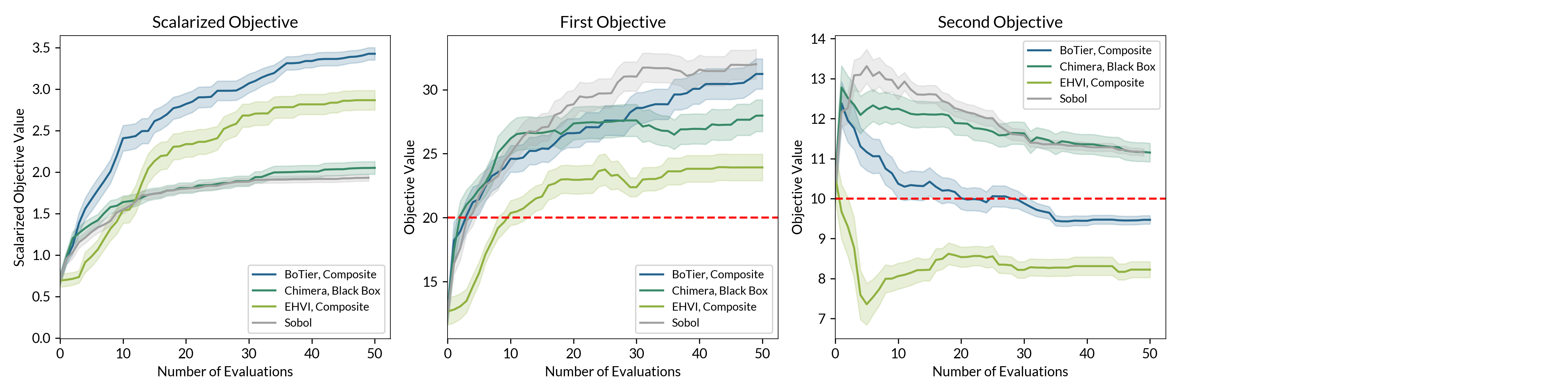}
 \caption{Comparison of different objective functions for multi-objective Bayesian Optimization: \textit{BoTier} (composite objective); \textit{Chimera} (black-box objective); and \textit{Expected Hypervolume Improvement}. All experiments were performed using  single-task Gaussian process model and the Expected Improvement acquisition function, if applicable. Experiments drawn from a Sobol sampler are included as a non-iterative, model-free baseline. All statistics were calculated over 50 independent runs, and intervals are plotted based on the standard error. Panel 1: Trajectories of the best hierarchical objective score obtained after $n$ experiments. Panel 2: Reaction yield at the best conditions found after $n$ experiments. Panel 3: Enzyme quantity at the best conditions found after $n$ experiments.}  
 \label{fig:alkox_em}
\end{figure}

\newpage

\subsection{Silver Nanoparticle Synthesis}

\begin{table}[H]
  \caption{Controllable variables for the Silver Nanoparticle Synthesis}.
  \begin{tabularx}{18cm}{cXccc}
  \toprule
  Index & Variable Description & Unit & Lower Bound & Upper Bound \\ \midrule \midrule
      0 & Flowrate of silver nitrate & $\mu$L/min & 4.5 & 42.8 \\
      1 & Flowrate of polyvinyl alcohol & $\mu$L/min & 10.0 & 40.0 \\
      2 & Flowrate of trisodium citrate & $\mu$L/min & 0.5 & 30.0 \\
      3 & Flowrate of seed solution & $\mu$L/min & 0.5 & 20 \\
      4 & Total reaction volume & $\mu$L/min & 200.0 & 1000.0 \\
    \bottomrule
  \end{tabularx}
  \label{tab:nanopart_inputs}
\end{table}

\begin{table}[H]
  \caption{Optimization objectives for the silver nanoparticle spectral composition optimization.}
  \centering
  \begin{tabularx}{18cm}{cccYc}
    \toprule
     & Objective & Direction & Calculation & Threshold \\ \midrule \midrule
    1 & Spectral Composition & $\uparrow$ & $y$ & 0.8 \\
    2 & Volume of seed solution & $\downarrow$ & $x_3$ & 9.0 \\
    \bottomrule
  \end{tabularx}
  \label{tab:nanopart_objectives}
\end{table}

\begin{figure}[H]    
 \centering
 \includegraphics[width=18cm]{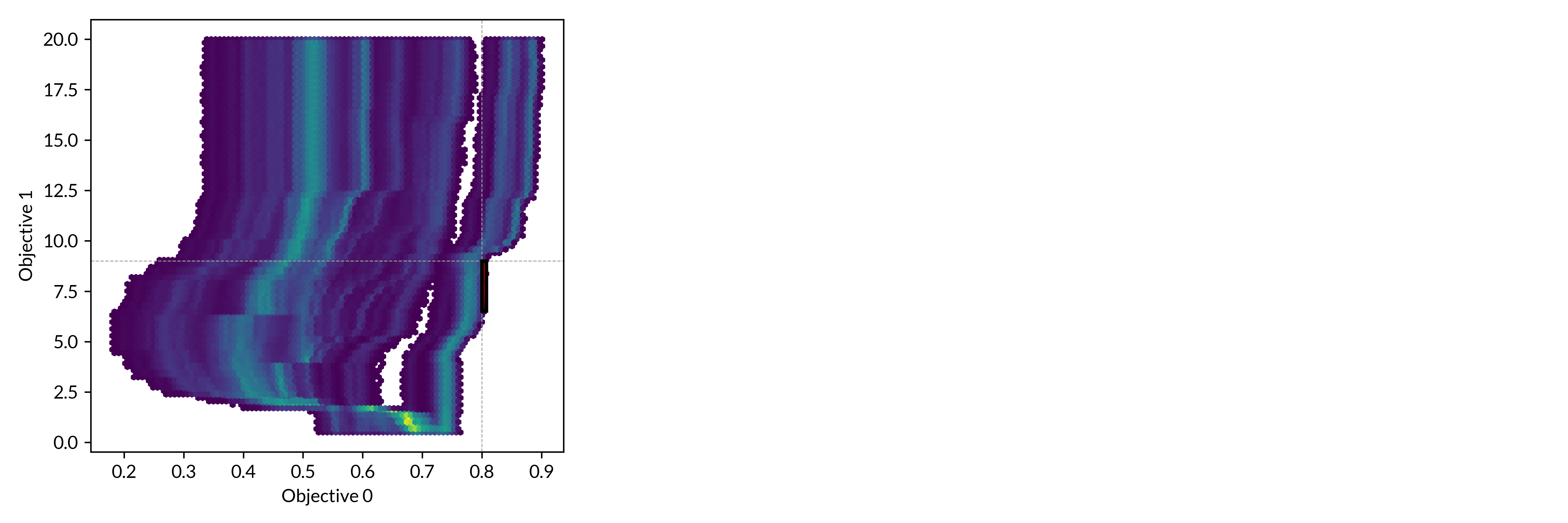}
 \caption{Objective density plots of the Silver Nanoparticle Spectral Composition Optimization, as described in Tab. \ref{tab:nanopart_inputs} and Tab. \ref{tab:nanopart_objectives}. The function was evaluated on a grid of $5\cdot10^6$ points drawn from a Sobol sampler. Data points that satisfy all objectives are shown in red.}
 \label{fig:nanopart_objectives}
\end{figure}

\begin{figure}[H]    
 \centering
 \includegraphics[width=18cm]{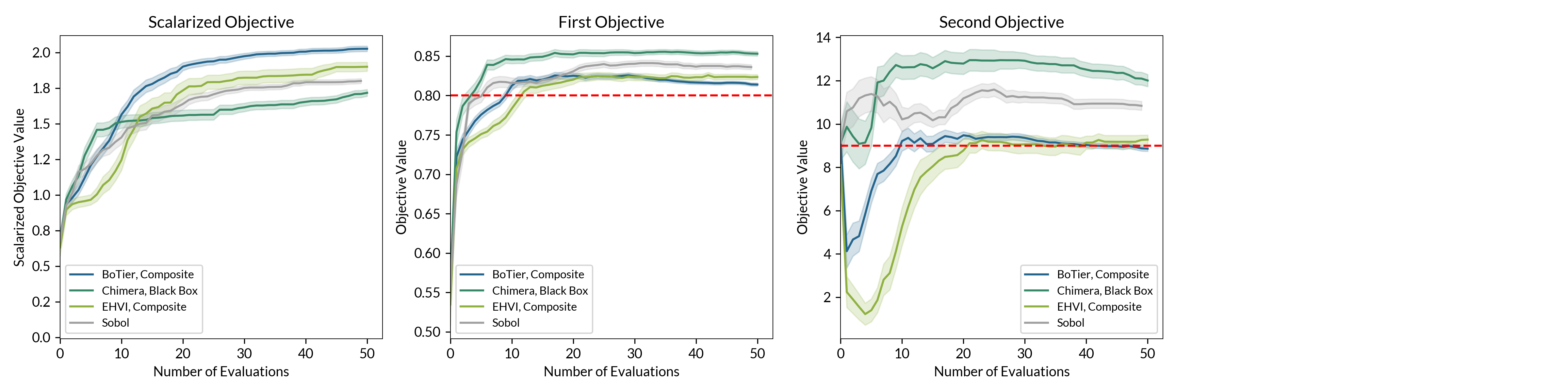}
 \caption{Comparison of different objective functions for multi-objective Bayesian Optimization: \textit{BoTier} (composite objective); \textit{Chimera} (black-box objective); and \textit{Expected Hypervolume Improvement}. All experiments were performed using  single-task Gaussian process model and the Expected Improvement acquisition function, if applicable. Experiments drawn from a Sobol sampler are included as a non-iterative, model-free baseline. All statistics were calculated over 50 independent runs, and intervals are plotted based on the standard error. Panel 1: Trajectories of the best hierarchical objective score obtained after $n$ experiments. Panel 2: Spectral composition at the best conditions found after $n$ experiments. Panel 3: Seed solution volume at the best conditions found after $n$ experiments.}  
 \label{fig:nanopart_em}
\end{figure}

\newpage

\section{Comparison of \textit{BoTier} and \textit{Chimera}}

To further evaluate the comparability of \textit{BoTier} and \textit{Chimera}, a systematic evaluation of the correlation between these two scalarization functions was performed. Each of the surfaces described in sections \ref{sec:si_benchmarks} and \ref{sec:si_benchmarks_emulated}, was evaluated on a grid of 10,000 data points drawn from a Sobol sampler. With these observations, the values of \textit{BoTier} and \textit{Chimera} were calculated for each point, and the Spearman rank correlation coefficient between both scores was computed.

\captionsetup{justification=centering}
\begin{table}[H]
\centering
\caption{Spearman $R$ between \textit{BoTier} and \textit{Chimera}, evaluated on a grid of 10,000 samples from each surface.}
\begin{tabular}{lr}
\toprule
\textbf{Surface}       & \textbf{Spearman $R$} \\ \midrule \midrule
BNH                    & 0.38311  \\
BNH*                   & 0.57663  \\
DH4                    & 0.99729 \\
DH4*                   & 0.99376 \\
DTLZ5                  & 0.97902 \\
DTLZ5*                 & 0.98092 \\
ZDT1                   & 0.99417 \\
ZDT1*                  & 0.99429 \\ 
Heterocyclic Suzuki–Miyaura Coupling & 0.99229 \\
Benzylation of $\alpha$-methylbenzylamine & 0.99844 \\
Enzymatic Alkoxylation & 0.99795 \\
Silver Nanoparticle Synthesis & 0.99057 \\

\bottomrule
\end{tabular}
\label{tab:spearman_table}
\end{table}

Overall, we see excellent rank correlation between \textit{BoTier} and \textit{Chimera}, further validating the design of \textit{BoTier}. The low degrees of correlation for the BNH surface can be attributed to a high density of function values that are close to the user-defined thresholds (see Fig. \ref{fig:bnh_a} and Fig. \ref{fig:bnh_b}), leading to numerical inconsistency between the implementations of \textit{BoTier} and \textit{Chimera}.

\end{si}

\end{document}